\documentclass[journal,10pt,twocolumn]{IEEEtran}

\usepackage{amsmath,amssymb,amsfonts}
\usepackage{algorithmic}
\usepackage{graphicx}
\usepackage{textcomp}
\usepackage{xcolor}
\usepackage{booktabs}
\usepackage{multirow}
\usepackage{url}
\usepackage{cite}
\usepackage{tikz}
\usetikzlibrary{shapes,arrows,positioning,calc,fit,backgrounds,trees,shadows}
\usepackage{array}
\usepackage{tabularx}
\usepackage{makecell}
\usepackage[colorlinks=true,linkcolor=primaryblue,citecolor=primaryblue,urlcolor=primaryblue]{hyperref}

\definecolor{primaryblue}{RGB}{24, 76, 120}
\definecolor{secondaryblue}{RGB}{225, 238, 248}
\definecolor{accentred}{RGB}{180, 40, 40}
\definecolor{accentgreen}{RGB}{34, 139, 34}
\definecolor{boxbg}{RGB}{250, 251, 252}
\definecolor{lightred}{RGB}{253, 237, 237}
\definecolor{lightgreen}{RGB}{235, 247, 238}

\begin{document}

\title{Trustworthy Agentic AI: A Comprehensive Cybersecurity and Systems Survey on Threat Landscapes, Defense Architectures, and Open Challenges}

\author{Seyedakbar~Mostafavi
\thanks{Seyedakbar Mostafavi is with the Department of Computer Engineering, Yazd University, Yazd 89195-741, Iran (e-mail: a.mostafavi@yazd.ac.ir).}%
\thanks{Manuscript received August 31, 2026. Corresponding author: Seyedakbar Mostafavi.}}

\maketitle

\begin{abstract}
The transition from passive foundation models to autonomous, goal-directed \textbf{Agentic AI systems} has introduced unprecedented capabilities by coupling recursive cognitive reasoning loops (e.g., ReAct, Tree-of-Thoughts, Reflexion), hierarchical memory architectures, live tool execution planes (Model Context Protocol, APIs, shell interpreters), and distributed multi-agent collaboration topologies. However, granting probabilistic neural reasoning cores execution authority across filesystems, networks, and cloud infrastructure dissolves classical security perimeters. In agentic systems, natural language simultaneously serves as input data, internal control code, and communication protocols, exposing a Turing-complete blast radius where untrusted data represents executable instructions.

This survey delivers the first comprehensive systems-security reference framework for \textbf{Trustworthy Agentic AI}, synthesizing 206 foundational studies and regulatory standards. We formalize the general agent architecture as a stateful 5-tuple and establish a \textbf{6-Dimensional Trustworthiness Taxonomy} encompassing \textit{Security, Safety \& Operational Robustness, Privacy, Explainability \& Verifiability, Fairness, and Accountability}. We systematically analyze full-spectrum threat surfaces across intra-execution loops (prompt injections, reasoning backdoors, hallucination cascades, tool parameter RCE, SSRF) and interaction planes (indirect prompt injections, vector database poisoning, embedding inversion, cascading swarm failures, Byzantine consensus subversion). To mitigate these threats, we formulate a multi-layered \textbf{Zero-Trust Defense-in-Depth Architecture} integrating Dual-LLM Inspector-Executor isolation, Capability-Based Access Control (CapBAC), kernel-level eBPF syscall probes, micro-VM sandboxing, signed vector embeddings, and Byzantine fault tolerance. Finally, we review standardized evaluation benchmarks (InjecAgent, AgentBench, CyberSecEval), map technical controls to international AI regulations (NIST AI RMF, EU AI Act, ISO/IEC 42001), and delineate high-priority open research frontiers.
\end{abstract}


\section{Introduction}\label{sec:intro}
The artificial intelligence landscape is undergoing a monumental paradigm shift from static foundation models that generate tokens within isolated conversational dialogs to \textbf{autonomous, goal-directed Agentic AI systems} \cite{xi2023rise, wang2024surveyagent, bommasani2021opportunities}. Unlike conversational large language models (LLMs) that operate as passive text processors, Agentic AI systems synthesize recursive cognitive reasoning loops (such as ReAct \cite{yao2022react}, Tree-of-Thoughts \cite{yao2023tree}, Graph-of-Thoughts \cite{besta2024graph}, and Reflexion \cite{shinn2023reflexion}), persistent multi-tier memory stores \cite{packer2023memgpt}, and live tool actuation planes \cite{schick2023toolformer, patil2023gorilla, anthropic2024mcp}. By autonomously decomposing abstract objectives, querying external databases, selecting APIs, executing shell commands, and collaborating within multi-agent networks \cite{wu2023autogen, hong2023metagpt}, agentic systems are revolutionizing software engineering, automated cyber defense, scientific discovery, and enterprise automation \cite{yang2024swe, deng2023mind2web}.

\begin{figure}[htbp]
\centering
\resizebox{\columnwidth}{!}{
\begin{tikzpicture}[node distance=0.8cm, auto, >=latex', thick]
    \tikzstyle{era} = [draw=primaryblue, fill=secondaryblue, rectangle, rounded corners, minimum height=2em, text width=8.5cm, align=center, font=\footnotesize\bfseries]
    \tikzstyle{line} = [draw=primaryblue, -latex', line width=1.2pt]

    \node [era] (e1) {1. Statistical ML Era (Pre-2018)\\Static Decision Boundaries, Supervised Classifiers};
    \node [era, below=0.4cm of e1] (e2) {2. Foundation Model Era (2018--2023)\\Passive Linguistic Synthesis, Stateless Chat Dialogs};
    \node [era, below=0.4cm of e2] (e3) {3. Agentic AI Era (2023--Present)\\Autonomous Agency, Persistent Memory, Live Tool Actuation};

    \path [line] (e1) -- (e2);
    \path [line] (e2) -- (e3);
\end{tikzpicture}
}
\caption{Historical Evolution of Artificial Intelligence Paradigms.}
\label{fig:ai_evolution}
\end{figure}
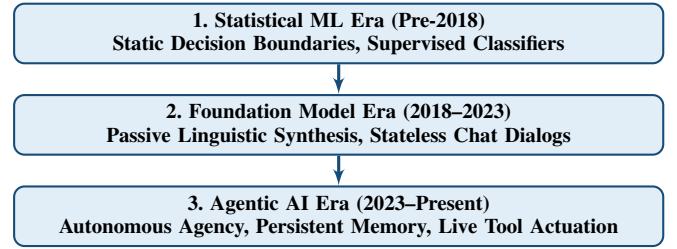

However, granting probabilistic neural engines autonomous execution authority across physical and digital environments shatters classical security perimeters \cite{schneier2023ai, moore2024autonomous}. In traditional software architectures, strict separation is enforced between untrusted input data and executable control logic, implemented through hardware-enforced Data Execution Prevention, $W \oplus X$ memory management policies, and strictly parameterized database interfaces \cite{saltzer1975protection, denning1976lattice}. In Agentic AI, this fundamental boundary collapses entirely: natural language simultaneously serves as the user input, the internal program logic, and the inter-component communication protocol:
\begin{equation}
\begin{split}
\text{Input Data} &\equiv \text{Control Logic} \\
&\equiv \text{Communication Protocol} = \text{Natural Language}
\end{split}
\label{eq:equivalence}
\end{equation}

Consequently, any untrusted data retrieved by an agent from a webpage, an email, an external database record, or a peer agent represents arbitrary, uncompiled executable code \cite{greshake2023not, zhan2024injecagent}. When an agent is empowered with tool execution capabilities, an adversarial input can trigger unauthorized remote code execution, sensitive data exfiltration, or cloud infrastructure compromise at machine speed \cite{perez2022ignore, willison2023prompt, abdelnabi2023not, rehberger2023pivoting}.

\subsection{Taxonomy of Agent Execution Archetypes}
Modern Agentic AI manifests across several distinct operational archetypes, each characterized by specialized execution environments, tool integration patterns, and operational failure surfaces. In software engineering, autonomous agents such as SWE-agent \cite{yang2024swe}, ChatDev \cite{qian2023chatdev}, and MetaGPT \cite{hong2023metagpt} ingest complex repository issue descriptions, navigate multi-file codebases, formulate reproduction scripts, edit source files, and execute compilation and test suites via interactive bash shells. These agents operate with broad filesystem read-write privileges, making any divergence in planning directly dangerous to repository integrity.

In interactive web environments, agents such as WebGPT \cite{nakano2021webgpt} and Mind2Web \cite{deng2023mind2web} parse Document Object Model (DOM) trees, formulate multi-step browser actions (including form submission, authentication handshakes, and page navigation), and interact with commercial and enterprise web portals. These systems operate in inherently untrusted execution contexts where adversarial third-party web content can easily poison the agent's perceptual stream. 

In cybersecurity operations, autonomous agents are increasingly deployed for dual-use missions, ranging from automated vulnerability discovery, penetration testing, and exploit chain synthesis to real-time security operations center (SOC) incident triage and automated patch verification \cite{bhatt2023purple, bhatt2024cyberseceval2, moore2024autonomous}. Concurrently, cyber-physical and embodied robotics agents translate high-level natural language intentions into low-level motor primitives for robotic arms, autonomous vehicles, and aerial drones (e.g., SayCan \cite{ahn2022saycan}, SurrealDriver \cite{jin2024surrealdriver}, VOYAGER \cite{wang2024voyager}). Finally, enterprise workflow orchestration swarms (such as AutoGen \cite{wu2023autogen}, CAMEL \cite{li2023camel}, and AgentVerse \cite{chen2023agentverse}) distribute complex organizational tasks across specialized agents, coordinating supply chain optimization, contract auditing, and multi-departmental business intelligence.

\subsection{The Cybersecurity Dilemma in Agentic Autonomy}
We formalize this fundamental security crisis as \textbf{The Agentic Cybersecurity Dilemma}, which emerges from three compounding structural tensions between agent capability and system controllability.

\textbf{Turing-Complete Blast Radius.} By equipping foundation models with interactive shell interpreters, persistent filesystem access, cloud APIs, and standardized Model Context Protocol (MCP) servers \cite{anthropic2024mcp, yang2024swe}, the blast radius of a successful prompt injection expands exponentially. Whereas attacking a conversational model results merely in toxic text generation, compromising an autonomous agent leads directly to arbitrary host takeover, remote code execution, and cloud privilege escalation across enterprise perimeters \cite{zhan2024injecagent, fang2024privilege}.

\textbf{Probabilistic and Non-Deterministic Control Flow.} Classical cybersecurity defenses rely upon deterministic state machines, static code analysis, and mathematically verifiable authorization invariants \cite{levy1984capability, rose2020zero}. In stark contrast, the cognitive core of an agent is inherently stochastic:
\begin{equation}
a_t \sim P(a_t \mid H_{t-1}, o_{t-1}; \theta)
\end{equation}
Minor semantic variations, sampling temperature fluctuations, or imperceptible prompt perturbations cause identical system states to yield radically different tool dispatch sequences, allowing malicious payloads to bypass static signature-based filters and runtime heuristic firewalls with high probability \cite{zou2023universal, tramer2020adaptive}.

\textbf{Dual-Use Machine-Speed Cyber Warfare.} Autonomous agents drastically compress the operational timeline of cyber engagements \cite{moore2024autonomous}. While defensive agents can automate security operations center triage, correlate distributed telemetry, and synthesize software patches in real time, adversaries can weaponize identical architectures to discover zero-day vulnerabilities, synthesize polymorphic exploits, and execute adaptive lateral movement at machine speed, creating an asymmetric offensive advantage \cite{bhatt2023purple, bhatt2024cyberseceval2, schneier2023ai}.

\begin{figure*}[htbp]
\centering
\resizebox{0.95\textwidth}{!}{
\begin{tikzpicture}[node distance=1.2cm, auto, >=latex', thick]
    \tikzstyle{agentnode} = [draw=primaryblue, fill=secondaryblue, rectangle, rounded corners, minimum height=3em, text width=3.8cm, align=center, font=\small\bfseries]
    \tikzstyle{gapnode} = [draw=accentred, fill=boxbg, rectangle, rounded corners, minimum height=3em, text width=4.5cm, align=center, font=\footnotesize]
    \tikzstyle{line} = [draw=primaryblue, -latex', line width=1.2pt]

    \node [agentnode] (agent) {Autonomous Agent\\$\mathcal{A} = \langle \mathcal{C}, \mathcal{M}, \mathcal{T}, \mathcal{E}, \Pi \rangle$};
    
    \node [gapnode, above left=0.6cm and 1.8cm of agent] (gap1) {\textbf{Gap 1: Input Unpredictability}\\Multi-step prompt injections, adversarial suffixes, jailbreaks};
    \node [gapnode, above right=0.6cm and 1.8cm of agent] (gap2) {\textbf{Gap 2: Internal Complexity}\\Reasoning drift, hallucination cascades, sleeper backdoors};
    \node [gapnode, below left=0.6cm and 1.8cm of agent] (gap3) {\textbf{Gap 3: Env. Variability}\\SSRF, sandbox breakout, resource exhaustion DoS};
    \node [gapnode, below right=0.6cm and 1.8cm of agent] (gap4) {\textbf{Gap 4: Untrusted Entities}\\RAG vector poisoning, Byzantine peer consensus subversion};

    \path [line] (gap1) -- (agent);
    \path [line] (gap2) -- (agent);
    \path [line] (gap3) -- (agent);
    \path [line] (gap4) -- (agent);
\end{tikzpicture}
}
\caption{The Four Critical Knowledge Gaps in Trustworthy Agentic AI Security.}
\label{fig:knowledge_gaps}
\end{figure*}
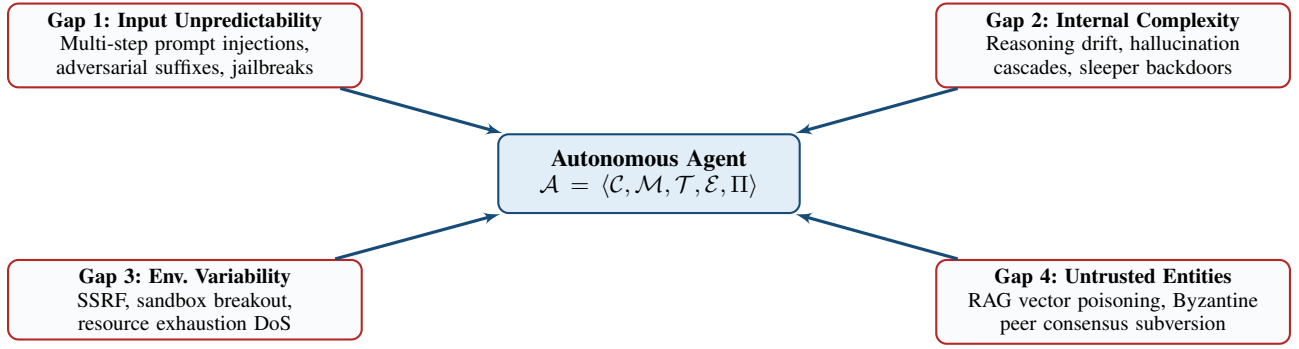

\subsection{Four Critical Knowledge Gaps in Agentic AI Security}
As illustrated in Fig.~\ref{fig:knowledge_gaps}, the vulnerabilities of modern Agentic AI stem from four foundational knowledge gaps that span the perception, reasoning, execution, and communication planes. 

The first gap concerns the \textbf{Unpredictability of Multi-Step User Inputs}, wherein multi-turn conversational interactions introduce subtle prompt drift, automated jailbreaks, and goal hijacking attacks that manipulate the agent into violating developer safety policies over extended trajectories \cite{chao2023jailbreaking, mehrotra2023tree, liu2023autodan, li2023deepinception, yong2023low, yuan2023gpt, jiang2024artprompt, levi2024vocabulary}. The second gap arises from the \textbf{Complexity of Internal Cognitive Reasoning}, where the agent's internal thought process constitutes a multi-branch graph structure rather than a simple feedforward pipeline. Minor hallucinations, semantic drift, or latent sleeper backdoors in intermediate thoughts compound multiplicatively into catastrophic tool execution failures \cite{besta2024graph, zhou2023language, ji2024testing, hubinger2024sleeper, xiang2024badchain, zhang2024how, mundler2024self}.

The third gap involves the \textbf{Variability of Operational Environments}, where deploying agents across diverse virtual filesystems, containerized networks, and cloud infrastructures introduces severe system-level vulnerabilities, including Server-Side Request Forgery (SSRF), sandbox breakout escapes, and algorithmic resource exhaustion denial-of-service \cite{agrawal2020firecracker, google2018gvisor, vieira2020fast, shumailov2021sponge, guastalla2023application}. Finally, the fourth gap emerges from \textbf{Interactions with Untrusted External Entities}, wherein interfacing with third-party web pages, external API providers, Model Context Protocol servers, persistent vector databases, and peer agents exposes the system to Indirect Prompt Injection (IPI), dense vector database poisoning, and Byzantine consensus subversion \cite{greshake2023not, zhan2024injecagent, zou2024poisonedrag, chen2024shadowcast, castro2002practical, li2024byzantine, gu2024cascading, zhang2024agentpoison}.

\subsection{Systematic Survey Methodology \& Corpus Identification}
To establish an exhaustive, evidence-backed foundation for this survey, we implemented a rigorous literature review protocol guided by the PRISMA-S standards for systematic literature syntheses \cite{page2021prisma}. Our bibliographic acquisition process spanned major computer science, cybersecurity, and artificial intelligence repositories, including IEEE Xplore, ACM Digital Library, USENIX Security, NDSS, IEEE S\&P, EuroS\&P, NeurIPS, ICML, ICLR, ACL, and arXiv.

Search queries integrated comprehensive boolean combinations across agent architectures, trustworthiness dimensions, and threat vectors:
\begin{multline}
\mathcal{Q} = (\text{Agentic AI} \lor \text{LLM Agent} \lor \text{Autonomous Agent}) \\
\land (\text{Trustworthy} \lor \text{Security} \lor \text{Prompt Injection} \\
\lor \text{Zero-Trust} \lor \text{Memory Poisoning} \lor \text{Byzantine Consensus})
\label{eq:prisma}
\end{multline}

Our rigorous multi-stage filtering and citation cross-validation protocol distilled a final reference corpus of \textbf{206 verified foundational publications}, spanning cognitive reasoning formalisms, systems sandboxing, vector memory integrity, and international regulatory governance.

\subsection{Comparison with Related Surveys and Our Contributions}
While several existing surveys explore general LLM applications \cite{bommasani2021opportunities, xi2023rise, wang2024surveyagent}, classical machine learning security \cite{goodfellow2014explaining, madry2018towards, papernot2016limitations}, or narrow prompt injection techniques \cite{zou2023universal, wei2023jailbroken}, they fail to provide a unified systems-security architecture for autonomous agent loops with persistent memory, tool execution, and multi-agent consensus. Table~\ref{tab:survey_comparison} provides a comprehensive comparison between our survey and related works in the literature.

\begin{table*}[htbp]
\caption{Comprehensive Comparison of This Survey with Existing Surveys in AI and Agent Security}
\label{tab:survey_comparison}
\centering
\scriptsize
\begin{tabularx}{\textwidth}{l c p{3.2cm} X p{4.2cm}}
\toprule
\textbf{Survey} & \textbf{Year} & \textbf{Primary Focus / Scope} & \textbf{Key Capabilities} & \textbf{Identified Limitations / Gaps} \\
\midrule
Xi et al. \cite{xi2023rise} & 2023 & General LLM agents & Cognitive architecture & Overlooks adversarial threat vectors and systems sandboxing. \\
Wang et al. \cite{wang2024surveyagent} & 2024 & Agent development & Planning \& evaluation & Lacks systems security and memory poisoning analysis. \\
Greshake et al. \cite{greshake2023not} & 2023 & Indirect prompt injection & App vulnerability demo & Limited to initial IPI discovery; lacks defense blueprint. \\
Zhan et al. \cite{zhan2024injecagent} & 2024 & Tool-integrated agents & Benchmarking across tools & Empirical attack suite; omits multi-agent and memory planes. \\
Deng et al. \cite{deng2023mind2web} & 2024 & Web interactive agents & Web task execution & Focuses on task success, not operational security. \\
Yao et al. \cite{yao2023tree} & 2023 & Deliberate planning & ToT reasoning search & Evaluates reasoning without adversarial threat models. \\
\textbf{Ours} & \textbf{2026} & \textbf{Trustworthy Agentic AI} & \textbf{6D Taxonomy, 4-Layer Zero-Trust, 206 Works} & \textbf{Unified zero-trust systems security blueprint with neural-symbolic verification synthesis.} \\
\bottomrule
\end{tabularx}
\end{table*}

\subsection{Formal Survey Research Questions (RQs)}
To establish an intellectually rigorous investigation into the trustworthiness of autonomous agentic systems, this survey is structured around five foundational Research Questions:

\textbf{RQ1: Foundational Systems-Level Formalization.} \textit{How can the dynamic closed-loop interaction among cognitive planning cores, hierarchical memory subsystems, live tool execution runtimes, and multi-agent network topologies be unified into a mathematically grounded, stateful systems security model?} We address this in Section~\ref{sec:arch} by formalizing the agent as a discrete-time 5-tuple $\mathcal{A} = \langle \mathcal{C}, \mathcal{M}, \mathcal{T}, \mathcal{E}, \Pi \rangle$ embedded within an adversarial POMDP framework.

\textbf{RQ2: Multi-Dimensional Trustworthiness Taxonomy.} \textit{What operational dimensions constitute trustworthiness in autonomous agentic ecosystems, and what mathematical formulations define their system invariants and cross-dimensional Pareto trade-offs?} We resolve this in Section~\ref{sec:taxonomy} through our 6-Dimensional Trustworthiness Taxonomy and cross-dimensional invariant analysis.

\textbf{RQ3: Full-Spectrum Vulnerability and Threat Modeling.} \textit{What are the exact attack mechanisms, adversary capabilities, and empirical failure modes across the intra-execution modules (perception, brain, action) and interaction surfaces (environment, persistent memory, peer swarms)?} We dissect these threat landscapes in Section~\ref{sec:intra} and Section~\ref{sec:interaction}.

\textbf{RQ4: Zero-Trust Defensive Systems Engineering.} \textit{How can classical systems security principles---including the Principle of Least Privilege, Capability-Based Access Control (CapBAC), kernel-level eBPF tracing, micro-VM sandboxing, cryptographic provenance, and Byzantine fault-tolerant consensus---be synthesized into an actionable Defense-in-Depth framework?} We formulate this multi-tiered architecture and its operational trade-offs in Section~\ref{sec:defense}.

\textbf{RQ5: Empirical Evaluation and Regulatory Compliance.} \textit{What empirical benchmarks and quantitative metrics assess agent trustworthiness, and how do technical architectural controls map to international regulatory mandates (such as the NIST AI RMF 1.0, EU AI Act, and ISO/IEC 42001)?} We synthesize the empirical benchmark landscape in Section~\ref{sec:benchmarks}, cross-regulatory enforcement in Section~\ref{sec:governance}, and chart strategic future frontiers in Section~\ref{sec:open}.

\subsection{Core Contributions of this Survey}
In addressing these foundational questions, this work delivers four principal intellectual contributions to the literature:

\textbf{1. Foundational Systems-Level Formalization and Boundary Collapse Framework.} We formalize the general architecture of autonomous agents as a stateful 5-tuple $\mathcal{A} = \langle \mathcal{C}, \mathcal{M}, \mathcal{T}, \mathcal{E}, \Pi \rangle$, modeling the closed-loop execution dynamics across cognitive planning, hierarchical memory stores, tool runtimes (including the Model Context Protocol), and multi-agent swarms. We mathematically couple this with an adversarial POMDP formulation that captures belief-state divergence under unvalidated environmental observations, establishing the Systems Boundary Collapse Principle where natural language simultaneously operates as untrusted data, executable control logic, and inter-process communication.

\textbf{2. Operational 6-Dimensional Trustworthiness Taxonomy and Pareto Dynamics.} Moving beyond static ML benchmark metrics, we formulate an operational, systems-level taxonomy spanning \textit{Security (CIA), Safety \& Operational Robustness, Privacy \& Data Protection, Explainability \& Verifiability, Fairness \& Non-Discrimination, and Accountability \& Provenance}. Each dimension is grounded in formal system invariants, empirical verification methods, and systems controls, accompanied by an explicit analysis of cross-dimensional Pareto tensions (security-latency, privacy-utility, and safety-autonomy).

\textbf{3. Full-Spectrum Dual-Plane Threat Dissection.} We provide an exhaustive, multi-tier threat analysis that deconstructs attack vectors across intra-execution planes (perception prompt injections, intermediate CoT reasoning poisoning, hallucination cascades, and tool parameter RCE/SSRF) and interaction planes (untrusted web environments, dense RAG vector poisoning, embedding inversion, Morris II generative worms, and Byzantine swarm subversion).

\textbf{4. Zero-Trust Defense Blueprint with Systems Trade-off Synthesis.} We formulate an actionable, multi-layered systems security blueprint synthesizing Dual-LLM Inspector-Executor isolation, Capability-Based Access Control (CapBAC), kernel-level eBPF syscall telemetry, micro-VM and WebAssembly sandboxing, cryptographically signed memory provenance, and Byzantine fault-tolerant consensus. Crucially, we provide a high-density comparative synthesis matrix analyzing the operational latency penalties, token amplification costs, computational footprints, and cloud/cluster deployability of each defensive mechanism.

\subsection{Paper Organization}
The remainder of this survey is organized as follows: Section~\ref{sec:arch} establishes the architectural foundations and execution workflows of Agentic AI. Section~\ref{sec:taxonomy} introduces the 6-Dimensional Trustworthiness Taxonomy. Section~\ref{sec:intra} examines intra-execution security across perception, brain, and action planes. Section~\ref{sec:interaction} analyzes interaction security across environments, memory stores, and multi-agent swarms. Section~\ref{sec:defense} presents the Zero-Trust Defense-in-Depth architecture. Section~\ref{sec:benchmarks} reviews empirical benchmarks, red-teaming frameworks, and quantitative metrics. Section~\ref{sec:governance} analyzes socio-technical governance and regulatory compliance (NIST AI RMF, EU AI Act, ISO/IEC 42001). Section~\ref{sec:open} delineates open research challenges and future research horizons. Finally, Section~\ref{sec:conclusion} concludes the paper. Table~\ref{tab:abbreviations} summarizes the key abbreviations used throughout the manuscript.

\begin{table}[htbp]
\caption{Summary of Key Abbreviations in Alphabetical Order}
\label{tab:abbreviations}
\centering
\scriptsize
\begin{tabularx}{\columnwidth}{l X}
\toprule
\textbf{Abbreviation} & \textbf{Definition} \\
\midrule
ACL & Agent Communications Language \\
ANN & Approximate Nearest Neighbor \\
ASR & Attack Success Rate \\
BFT / PBFT & Byzantine Fault Tolerance / Practical BFT \\
BRI & Blast Radius Index \\
CapBAC & Capability-Based Access Control \\
CFP & Cascade Failure Probability \\
CoT / ToT / GoT & Chain-of-Thought / Tree-of-Thoughts / Graph-of-Thoughts \\
DID & Decentralized Identifier (W3C Standard) \\
DPI / IPI & Direct Prompt Injection / Indirect Prompt Injection \\
eBPF & Extended Berkeley Packet Filter \\
FSRR & False Safety Rejection Rate \\
HIC / HITL / HOTL & Human-in-Command / Human-in-the-Loop / Human-on-the-Loop \\
HNSW & Hierarchical Navigable Small World \\
MCP & Model Context Protocol \\
MCTS & Monte Carlo Tree Search \\
MIA & Membership Inference Attack \\
PoLP & Principle of Least Privilege \\
RAG & Retrieval-Augmented Generation \\
RCE & Remote Code Execution \\
SMT & Satisfiability Modulo Theories \\
SSRF & Server-Side Request Forgery \\
TEE & Trusted Execution Environment \\
UPR & Utility Preservation Rate \\
\bottomrule
\end{tabularx}
\end{table}

\section{Foundational Architecture of Agentic AI}\label{sec:arch}

\subsection{Unified Conceptual Framework and Formal System Tuple}
To systematically analyze vulnerabilities, attack surfaces, and defensive architectures in autonomous agent systems, we formalize an Agentic AI system as a stateful, discrete-time autonomous 5-tuple operating over execution steps $t \in \{1, 2, \dots, T\}$ within an operational environment $\mathcal{E}$ \cite{xi2023rise, wang2024surveyagent}:
\begin{equation}
\mathcal{A} = \langle \mathcal{C}, \mathcal{M}, \mathcal{T}, \mathcal{E}, \Pi \rangle
\end{equation}

Within this formalization, $\mathcal{C}$ denotes the \textbf{Cognitive Reasoning Core} parameterized by foundation model weights $\theta$ (e.g., GPT-4 \cite{achiam2023gpt4}, Gemini \cite{team2023gemini}, LLaMA 3 \cite{dubey2024llama3}, Mistral \cite{jiang2023mistral}, Claude 3 \cite{anthropic2024claude3}), responsible for goal decomposition, intermediate reasoning, and decision synthesis. The \textbf{Hierarchical Memory Subsystem}, denoted $\mathcal{M} = \{\mathcal{M}_{\text{work}}, \mathcal{M}_{\text{episodic}}, \mathcal{M}_{\text{semantic}}\}$, manages information persistence across ephemeral attention windows, dense vector indices, and structured parametric knowledge stores \cite{packer2023memgpt}. 

The \textbf{Tool and Action Plane}, $\mathcal{T} = \{t_1, t_2, \dots, t_k\}$, defines the execution interface through which the agent causes external side effects. Each tool $t_j = \langle \text{Schema}_j, \text{Exec}_j, \text{Perm}_j \rangle$ is defined by a formal parameter schema, an underlying execution runtime, and a security permission envelope \cite{schick2023toolformer, patil2023gorilla, anthropic2024mcp}. The \textbf{External Operational Environment} $\mathcal{E}$ encompasses filesystems, network sockets, external databases, cloud APIs, human operators, and peer agents. Finally, $\Pi$ represents the \textbf{Agent Execution Policy or Orchestration Protocol}, which governs how historical interaction traces and memory states are mapped onto next-step thoughts, reflections, and tool invocations \cite{khattab2023dspy}.

\subsection{Partially Observable Markov Decision Process (POMDP) Formulation}
In real-world deployments, an autonomous agent does not possess direct access to the complete, underlying system state $s_t \in \mathcal{S}$. Instead, agent execution operates under partial observability, modeled formally as a Partially Observable Markov Decision Process (POMDP) defined by the 7-tuple $\langle \mathcal{S}, \mathcal{A}_{\text{act}}, \mathcal{T}_{\text{trans}}, \mathcal{R}, \Omega, \mathcal{O}, \gamma \rangle$. Here, $\mathcal{S}$ represents the set of hidden ground-truth system states (including full operating system memory, remote database tables, and cloud infrastructure states); $\mathcal{A}_{\text{act}}$ represents the discrete or structured action space exposed via tool plane $\mathcal{T}$; $\mathcal{T}_{\text{trans}}(s_{t+1} \mid s_t, a_t)$ denotes the state transition probability distribution of the environment; $\mathcal{R}(s_t, a_t)$ is the reward function quantifying goal fulfillment; $\Omega$ is the set of discrete observations emitted by the environment (such as stdout text, JSON payloads, HTTP responses, or error codes); $\mathcal{O}(o_{t+1} \mid s_{t+1}, a_t)$ is the observation probability distribution; and $\gamma \in [0, 1)$ is the temporal discount factor.

\textbf{Belief State Transition Dynamics.} Because the true state $s_t$ is unobservable, the agent maintains an internal belief state $b_t(s) = \Pr(s_t = s \mid H_t)$, representing a probability distribution over states conditioned on the execution history $H_t = (o_0, a_0, o_1, a_1, \dots, o_t)$. Upon executing action $a_t$ and receiving a new environmental observation $o_{t+1}$, the agent's belief state evolves according to the recursive Bayesian belief update:
\begin{equation}
\begin{split}
&b_{t+1}(s') = \\
&\frac{\mathcal{O}(o_{t+1} \mid s', a_t) \sum_{s \in \mathcal{S}} \mathcal{T}_{\text{trans}}(s' \mid s, a_t) b_t(s)}{\sum_{s'' \in \mathcal{S}} \mathcal{O}(o_{t+1} \mid s'', a_t) \sum_{s \in \mathcal{S}} \mathcal{T}_{\text{trans}}(s'' \mid s, a_t) b_t(s)}
\end{split}
\label{eq:belief_update}
\end{equation}
The agent's policy $\Pi(a_t \mid b_t)$ maps this internal belief state into a tool invocation $a_t \in \mathcal{A}_{\text{act}}$ to maximize expected cumulative discounted return:
\begin{equation}
\Pi^* = \arg\max_\Pi \mathbb{E}\left[\sum_{t=0}^T \gamma^t \mathcal{R}(s_t, a_t) \;\middle|\; s_0 \sim b_0, a_t \sim \Pi(\cdot \mid b_t)\right]
\label{eq:pomdp_objective}
\end{equation}

\textbf{Adversarial Belief Divergence.} Security vulnerabilities emerge precisely when untrusted external inputs or poisoned observations $o_{\text{adv}} \in \Omega$ corrupt the observation likelihood $\mathcal{O}(o_{\text{adv}} \mid s', a_t)$. An adversary injects malicious contextual observations to induce maximum divergence between the true environmental state distribution and the agent's internal belief distribution $b_t^{\text{adv}}$:
\begin{equation}
\begin{split}
&D_{\text{KL}}\left(b_t^{\text{true}}(s) \parallel b_t^{\text{adv}}(s)\right) \gg 0 \implies \\
&\arg\max_{a} \mathbb{E}_{s \sim b_t^{\text{adv}}}[\mathcal{R}(s, a)] \neq \arg\max_{a} \mathbb{E}_{s \sim b_t^{\text{true}}}[\mathcal{R}(s, a)]
\end{split}
\label{eq:adversarial_divergence}
\end{equation}
This divergence steers the policy $\Pi(a_t \mid b_t^{\text{adv}})$ toward catastrophic, unauthorized, or policy-violating tool actions while maintaining internal model confidence.

\begin{figure*}[htbp]
\centering
\resizebox{0.95\textwidth}{!}{
\begin{tikzpicture}[node distance=1.1cm, auto, >=latex', thick]
    \tikzstyle{layerbox} = [draw=primaryblue, fill=boxbg, rectangle, rounded corners, inner sep=8pt, text width=14.5cm]
    \tikzstyle{subblock} = [draw=primaryblue, fill=secondaryblue, rectangle, rounded corners, minimum height=2.2em, text width=3.1cm, align=center, font=\footnotesize\bfseries]
    \tikzstyle{databox} = [draw=accentred, fill=white, rectangle, rounded corners, minimum height=2em, text width=4.3cm, align=center, font=\footnotesize]
    \tikzstyle{line} = [draw=primaryblue, -latex', line width=1.2pt]

    \node [layerbox] (perception) {
        \textbf{1. Perception Layer (Input Formatting \& Context Ingestion)}\\[1mm]
        \begin{tikzpicture}
            \node [subblock, text width=4.3cm] (p1) at (0,0) {User Prompt\\($x_{\text{user}}$)};
            \node [subblock, text width=4.3cm] (p2) at (4.8,0) {Observations ($o_{t-1}$)\\DOM, Images};
            \node [subblock, text width=4.3cm] (p3) at (9.6,0) {Memory Context\\RAG Documents};
        \end{tikzpicture}
    };

    \node [layerbox, below=0.6cm of perception] (brain) {
        \textbf{2. Cognitive Reasoning Core ($\mathcal{C}$) --- The Brain}\\[1mm]
        \begin{tikzpicture}
            \node [subblock, text width=3.1cm] (b1) at (0,0) {Goal Decomposition\\Hierarchical Tasks};
            \node [subblock, text width=3.1cm] (b2) at (3.5,0) {Planning Loops\\ReAct, ToT, GoT};
            \node [subblock, text width=3.1cm] (b3) at (7.0,0) {Self-Reflection\\Reflexion, Refine};
            \node [subblock, text width=3.1cm] (b4) at (10.5,0) {Decision Making\\Tool JSON Schema};
        \end{tikzpicture}
    };

    \node [layerbox, below=0.6cm of brain] (action) {
        \textbf{3. Tool \& Action Plane ($\mathcal{T}$) --- Actuation}\\[1mm]
        \begin{tikzpicture}
            \node [subblock, text width=3.1cm] (a1) at (0,0) {Model Context Protocol\\(MCP Client)};
            \node [subblock, text width=3.1cm] (a2) at (3.5,0) {OS Command Shells\\(Bash, PowerShell)};
            \node [subblock, text width=3.1cm] (a3) at (7.0,0) {Web Automation\\(Playwright)};
            \node [subblock, text width=3.1cm] (a4) at (10.5,0) {Cloud APIs\\(S3, SQL, REST)};
        \end{tikzpicture}
    };

    \node [layerbox, below=0.6cm of action] (environment) {
        \textbf{4. Operational Environment ($\mathcal{E}$) \& Multi-Agent Swarm Ecosystem}\\[1mm]
        \begin{tikzpicture}
            \node [databox, text width=4.3cm] (e1) at (0,0) {Filesystems \& Cloud};
            \node [databox, text width=4.3cm] (e2) at (4.8,0) {Peer Agents (AutoGen, MetaGPT)};
            \node [databox, text width=4.3cm] (e3) at (9.6,0) {Human Operators (HITL)};
        \end{tikzpicture}
    };

    \path [line] (perception) -- (brain);
    \path [line] (brain) -- (action);
    \path [line] (action) -- (environment);
    \draw [line] (environment.west) -- ++(-0.4,0) |- (perception.west) node[pos=0.25, above, rotate=90, font=\footnotesize\bfseries] {Feedback ($o_t \leftarrow \mathcal{E}(a_t)$)};
\end{tikzpicture}
}
\caption{General Architecture and Closed-Loop Execution Workflow of Agentic AI Systems.}
\label{fig:general_architecture}
\end{figure*}
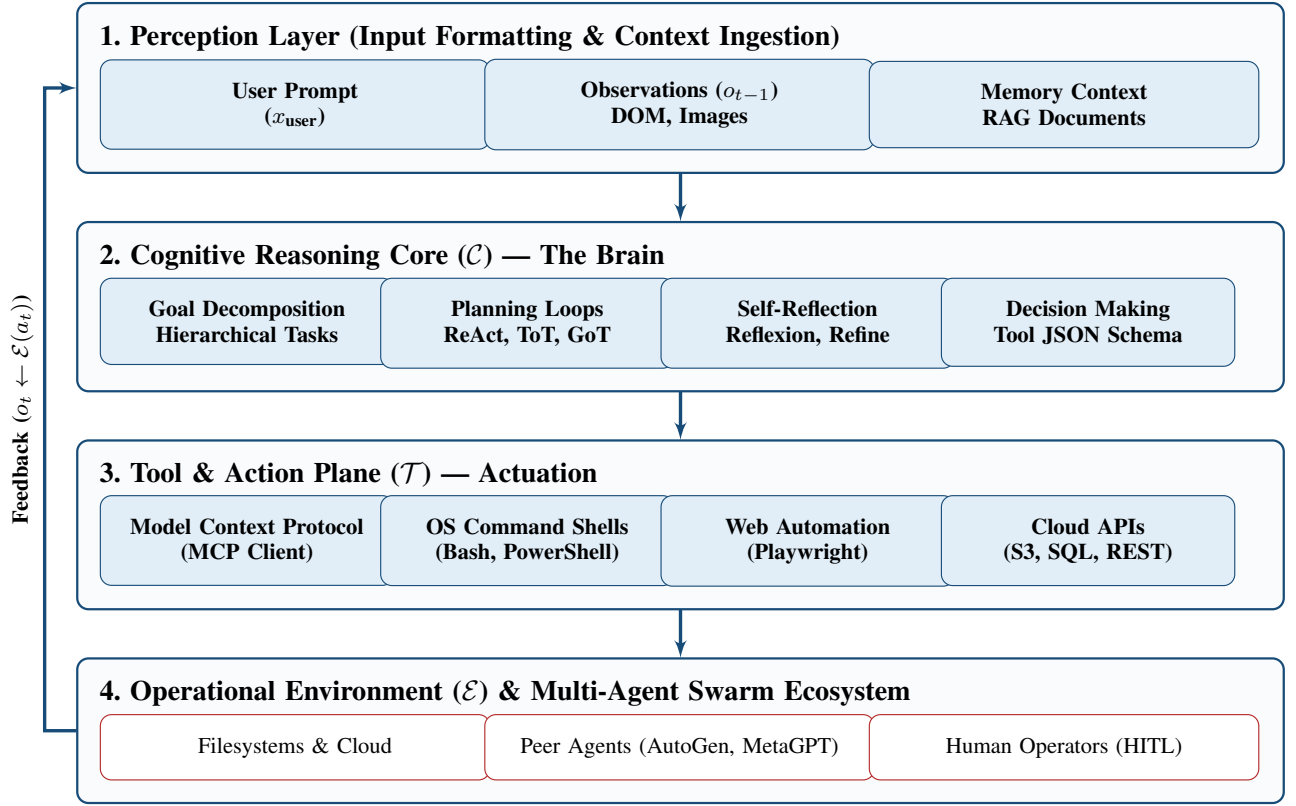

\subsection{Cognitive Reasoning \& Planning Paradigms}
The cognitive core $\mathcal{C}$ decomposes complex, high-level objectives into actionable execution sequences using structured reasoning paradigms that govern how the model reasons, samples actions, and incorporates environmental feedback.

\textbf{Linear and Interleaved Reasoning (CoT and ReAct).} In basic sequential architectures, Chain-of-Thought (CoT) prompting encourages the model to generate intermediate reasoning tokens before emitting a final decision \cite{wei2022chain, kojima2022large}. However, pure CoT operates open-loop without environmental grounding. To enable closed-loop execution, the ReAct paradigm interleaves reasoning traces ($th_t$) and action dispatches ($a_t$) dynamically \cite{yao2022react}. At execution step $t$, given the prior trajectory history $H_{t-1} = (th_1, a_1, o_1, \dots, th_{t-1}, a_{t-1}, o_{t-1})$ and current observation $o_{t-1}$, the agent executes:
\begin{align}
th_t &\sim \mathcal{C}(H_{t-1}, o_{t-1}; \theta) \\
a_t &\sim \mathcal{C}(H_{t-1}, th_t; \theta) \\
o_t &\leftarrow \mathcal{E}(a_t)
\end{align}
This interleaved cycle allows the agent to update its planning trajectory dynamically based on real-time feedback, error codes, and tool outputs.

\textbf{Search-Based Deliberate Planning (ToT, GoT, and LATS).} To tackle complex combinatorial problem domains, deliberate planning generalizes linear thought sequences into search graphs $\mathcal{G} = (\mathcal{V}, \mathcal{E})$ \cite{yao2023tree, besta2024graph, zhou2023language}. Under the Tree-of-Thoughts (ToT) and Graph-of-Thoughts (GoT) frameworks, intermediate cognitive states $v \in \mathcal{V}$ are evaluated by heuristic evaluation functions $V(v) = \mathbb{E}_{\mathcal{C}}[\text{Task Success} \mid v]$ using Breadth-First Search (BFS), Depth-First Search (DFS), or Monte Carlo Tree Search (MCTS) algorithms with lookahead rollouts \cite{wang2023plan}. While search-based planning dramatically improves task success in mathematical and software engineering domains, it exponentially expands the attack surface, providing adversarial inputs multiple candidate pathways to bias the search tree.

\textbf{Verbal Reinforcement Learning and Episodic Self-Correction.} To facilitate iterative improvement without parameter fine-tuning, frameworks such as Reflexion \cite{shinn2023reflexion} and Self-Refine \cite{madaan2023self} equip agents with verbal self-critique mechanisms. When an action trajectory fails validation checks or unit tests, an auxiliary evaluation model generates a natural language reflection $r_t \sim \mathcal{C}(H_t, \text{Feedback})$, which is stored in episodic memory to steer subsequent planning iterations. In embodied environments, grounded planning systems (such as VOYAGER \cite{wang2024voyager}, DEPS \cite{wang2023deps}, and SayCan \cite{ahn2022saycan}) extend this principle by maintaining evolving skill libraries and automated curricula grounded in physical affordances.

\subsection{Hierarchical Memory Subsystems}
To overcome the finite context window limits of foundation models and maintain persistent behavioral identity over long horizons, agents implement multi-tiered memory architectures \cite{packer2023memgpt}.

\textbf{Working Context Memory ($\mathcal{M}_{\text{work}}$).} Working memory corresponds to the active attention context window holding immediate system instructions, few-shot demonstration exemplars, active tool schemas, and recent conversation turns \cite{vaswani2017attention, ding2024longrope}. While modern models support extensive context windows, attention degradation over long token spans (the ``lost in the middle'' phenomenon) creates operational vulnerabilities.

\textbf{Episodic Vector Memory ($\mathcal{M}_{\text{episodic}}$).} Episodic memory captures historical execution logs, conversation histories, and external documents. These records are projected into dense semantic vector spaces $\mathbb{R}^d$ using neural embedding models and indexed via Hierarchical Navigable Small World (HNSW) \cite{malkov2018efficient} or FAISS \cite{johnson2019billion} graph structures. When the agent formulates a query $\mathbf{q}$, relevant memory chunks are retrieved via top-$k$ cosine similarity:
\begin{equation}
\text{Retrieve}(\mathbf{q}, k) = \arg\max_{m_1, \dots, m_k \in \mathcal{M}_{\text{episodic}}} \sum_{i=1}^k \frac{\mathbf{E}(\mathbf{q}) \cdot \mathbf{E}(m_i)}{\|\mathbf{E}(\mathbf{q})\|_2 \|\mathbf{E}(m_i)\|_2}
\label{eq:retrieval}
\end{equation}

\textbf{Semantic Knowledge and Virtual Context Paging.} Semantic and procedural memory $\mathcal{M}_{\text{semantic}}$ encompasses both the parametric knowledge internalized in model weights $\theta$ and structured Knowledge Graphs (KGs) that encode domain-specific rules and tool hierarchies \cite{patil2023gorilla, schick2023toolformer, pan2024survey}. To seamlessly bridge fast working memory and large-scale external stores, operating-system-inspired architectures like MemGPT \cite{packer2023memgpt} implement virtual context paging, enabling agents to dynamically swap memory blocks between context tiers via explicit function calls.

\subsection{Tool Integration Plane \& Model Context Protocol (MCP)}
The Action Plane $\mathcal{T}$ transforms abstract reasoning into real-world computational side effects through structured tool invocation paradigms.

\textbf{Structured Function Calling and Schema Validation.} Modern foundation models are fine-tuned to emit structured JSON or XML payloads adhering to strict JSON Schema specifications \cite{patil2023gorilla, shen2023hugginggpt}. The agent runtime parses these emitted payloads, validates argument types, and dispatches requests to local system binaries or remote web services.

\textbf{The Model Context Protocol (MCP) Standard and Transport Layers.} In late 2024, the Model Context Protocol (MCP) emerged as an open client-server JSON-RPC 2.0 specification standardizing how AI agents discover tools, access secure resources, and execute operations across local and distributed environments \cite{anthropic2024mcp}. MCP establishes three foundational primitives: \textit{Prompts} (pre-engineered contextual workflows), \textit{Resources} (structured, read-only contextual data), and \textit{Tools} (executable functions with real-world side effects). Operationally, MCP supports two primary communication transports: local process pipes over standard input/output (\texttt{stdio}) for single-host tool execution, and HTTP with Server-Sent Events (\texttt{SSE}) for remote, distributed microservice invocation. While MCP standardizes tool interoperability across heterogeneous environments, it also creates an expansive attack surface: rogue tool servers can publish poisoned function schemas that execute prompt injection at registration time, or manipulate tool response payloads to hijack downstream agent planning \cite{zhang2024supplychain}.

\textbf{Distributed Cluster Execution and Runtime Sandboxes.} In enterprise and high-performance computing clusters, autonomous agents are scheduled across distributed compute nodes using distributed orchestration runtimes such as Ray, Celery, and Kubernetes. Because agents execute untrusted code and arbitrary commands, contemporary cluster deployments enforce multi-tiered isolation. These range from containerized Linux namespaces (such as SWE-agent \cite{yang2024swe}) and headless browser automation engines (such as Playwright, Mind2Web \cite{deng2023mind2web}, and WebGPT \cite{nakano2021webgpt}) to hardware-virtualized micro-virtual machines (such as AWS Firecracker \cite{agrawal2020firecracker} and Google gVisor \cite{google2018gvisor}). Enforcing deterministic boundary isolation across distributed agent worker pools remains a vital systems requirement for cluster deployments.

\subsection{Multi-Agent Collaboration Topologies}
To resolve large-scale, complex enterprise workflows, individual agents are orchestrated into distributed multi-agent networks operating across diverse topological structures (Table~\ref{tab:agent_frameworks}).

\begin{table*}[htbp]
\caption{Comparison of Leading Agent Frameworks and Architectures}
\label{tab:agent_frameworks}
\centering
\scriptsize
\begin{tabularx}{\textwidth}{l p{3.2cm} p{2.6cm} p{3.2cm} X}
\toprule
\textbf{Framework} & \textbf{Primary Domain} & \textbf{Reasoning} & \textbf{Memory Model} & \textbf{Key Vulnerability Point} \\
\midrule
AutoGPT \cite{xi2023rise} & General automation & ReAct & Working + Vector DB & Unsanitized tool execution; loops. \\
AutoGen \cite{wu2023autogen} & Multi-agent chat & Conversable & Context passing & Byzantine peer injection; no signing. \\
MetaGPT \cite{hong2023metagpt} & Software engineering & SOP role play & Pub/Sub shared & Cascading hallucination propagation. \\
ChatDev \cite{qian2023chatdev} & Collaborative dev & Waterfall & Role buffers & Cross-role prompt injection via code. \\
SWE-agent \cite{yang2024swe} & Issue resolution & ACI + ReAct & Working + diffs & Shell parameter injection in bash. \\
MemGPT \cite{packer2023memgpt} & Conversational OS & Context paging & Multi-tier vector & Vector poisoning \& inversion. \\
\bottomrule
\end{tabularx}
\end{table*}

\textbf{Hierarchical and Communicative Orchestration.} In hierarchical orchestration topologies, such as MetaGPT \cite{hong2023metagpt} and AutoGen \cite{wu2023autogen}, a centralized supervisor agent receives high-level goals and delegates specific sub-tasks to subordinate specialized agents according to Standard Operating Procedures (SOPs). Conversely, in peer-to-peer communicative meshes (e.g., CAMEL \cite{li2023camel}, AgentVerse \cite{chen2023agentverse}, and DyLAN \cite{liu2023dylan}), agents negotiate directly through conversational turn-taking and contract-net protocols without centralized bottlenecks.

\textbf{Shared Blackboard Swarms and Game-Theoretic Consensus.} In blackboard-oriented architectures, such as ChatDev \cite{qian2023chatdev} and Generative Agents \cite{park2023generative}, multiple agents communicate asynchronously by reading from and writing to a shared global state repository. To improve reasoning fidelity, multi-agent debate frameworks employ game-theoretic cross-examination, iterative voting, and consensus mechanisms \cite{nisan2007algorithmic, sandholm2002algorithm, chan2024chateval}. However, as analyzed in Section~\ref{sec:interaction}, these collaborative topologies introduce critical vulnerabilities, including cascading hallucination amplification, AI worm propagation, and Byzantine consensus subversion.

\section{The 6-Dimensional Trustworthiness Taxonomy}\label{sec:taxonomy}

In classical machine learning, trustworthiness has traditionally been evaluated as a static statistical property over fixed benchmark datasets, measured through test-set accuracy, out-of-distribution generalization metrics, and bounded $\ell_p$-norm adversarial robustness \cite{goodfellow2014explaining, madry2018towards, papernot2016limitations, carlini2017towards}. In the context of \textbf{Agentic AI}, however, trustworthiness must be fundamentally reformulated as a \textbf{dynamic, systems-level operational guarantee}. Because autonomous agents possess execution agency, maintain persistent state across memory tiers, and generate real-world side effects through live tool dispatches, an operational failure in any single cognitive or execution module compromises the integrity of the entire socio-technical system \cite{amodei2016concrete, hendrycks2021unsolved}.

To establish an overarching conceptual structure for analyzing and evaluating agent trustworthiness, we formalize the \textbf{6-Dimensional Trustworthiness Taxonomy} for Agentic AI Systems (Fig.~\ref{fig:taxonomy_tree}).

\begin{figure*}[htbp]
\centering
\resizebox{0.95\textwidth}{!}{
\begin{tikzpicture}[node distance=0.8cm, auto, >=latex', thick]
    \tikzstyle{card} = [draw=primaryblue, fill=boxbg, rectangle, rounded corners, minimum height=4.2em, text width=4.5cm, align=left, font=\scriptsize]

    \node [card] (c1) {
        \textbf{1. Security (Sec. 3.1)}\\
        $\bullet$ Confidentiality: System prompt shielding\\
        $\bullet$ Integrity: Memory \& tool validation\\
        $\bullet$ Availability: DoS \& recursion defense
    };
    \node [card, right=0.5cm of c1] (c2) {
        \textbf{2. Safety \& Robustness (Sec. 3.2)}\\
        $\bullet$ Bounded Action Envelopes \& rollback\\
        $\bullet$ Hallucination Suppression in tools\\
        $\bullet$ OOD Observation Fault Tolerance
    };
    \node [card, right=0.5cm of c2] (c3) {
        \textbf{3. Privacy \& Data Protection (Sec. 3.3)}\\
        $\bullet$ Zero-Leakage API Tool Dispatches\\
        $\bullet$ Embedding Inversion Resistance\\
        $\bullet$ Multi-Tenant Context Hygiene
    };

    \node [card, below=0.5cm of c1] (c4) {
        \textbf{4. Explainability \& Verifiability (Sec. 3.4)}\\
        $\bullet$ Action-Chain Audit Traceability\\
        $\bullet$ Neural-Symbolic SMT Verifiers\\
        $\bullet$ Faithfulness of Reasoning Chains
    };
    \node [card, below=0.5cm of c2] (c5) {
        \textbf{5. Fairness \& Non-Discrimination (Sec. 3.5)}\\
        $\bullet$ Balanced Tool Priority Scheduling\\
        $\bullet$ Anti-Collusion in Multi-Agent Bidding\\
        $\bullet$ Demographic Parity in Decisions
    };
    \node [card, below=0.5cm of c3] (c6) {
        \textbf{6. Accountability \& Provenance (Sec. 3.6)}\\
        $\bullet$ Cryptographic Action Signatures (DIDs)\\
        $\bullet$ Tamper-Evident Merkle Logs\\
        $\bullet$ Legally Binding Agent Non-Repudiation
    };
\end{tikzpicture}
}
\caption{The 6-Dimensional Trustworthiness Taxonomy for Agentic AI Systems.}
\label{fig:taxonomy_tree}
\end{figure*}

\subsection{Dimension 1: Security (Confidentiality, Integrity, Availability)}
Security governs the resilience of the agent's cognitive core, memory subsystems, tool execution planes, and inter-agent communication channels against intentional, malicious adversarial attacks \cite{saltzer1975protection, denning1976lattice}. In agentic architectures, classical information security goals manifest through specialized operational invariants.

\textbf{Confidentiality of Internal State and Secrets.} Confidentiality requires ensuring that sensitive system prompts, proprietary tool schemas, API access credentials, and retrieved private episodic memories are never exfiltrated or disclosed via tool payloads or adversarial prompt manipulation \cite{zhang2024effective, geiping2024coercing, agarwal2024investigating}. Formally, the information-theoretic advantage of an adversary $\mathcal{A}_{\text{adv}}$ in recovering secret credentials or hidden system prompts $k_{\text{secret}} \in \mathcal{K}$ over security parameter $\lambda$ must be strictly bounded:
\begin{equation}
\mathbf{Adv}_{\mathcal{A}}^{\text{leak}}(\lambda) = \left| \Pr\left(\mathcal{A}_{\text{adv}}(x_{\text{adv}}) = k_{\text{secret}}\right) - \frac{1}{|\mathcal{K}|} \right| \le \operatorname{negl}(\lambda)
\end{equation}
In practice, achieving this requires strict output boundary filtering, parameter redaction, and prompt-shielding mechanisms that isolate internal system instructions from untrusted data channels.

\textbf{Integrity of Planning and Tool Invocations.} Integrity guarantees that the agent's internal reasoning loop, intermediate planning trees, memory embeddings, and tool dispatch parameters remain unaltered by unauthorized external entities or poisoned data sources \cite{xiang2024badchain, zou2024poisonedrag}. If an attacker can manipulate intermediate reasoning steps ($th_t$) or inject untrusted data into memory stores, the execution policy will diverge from developer intent. Preserving integrity requires cryptographic verification of external data sources and capability-based authorization gates for all tool calls.

\textbf{Availability and Computational Resilience.} Availability guarantees continuous, deterministic service operation, protecting agent systems against computational denial-of-service (DoS) attacks \cite{guastalla2023application}. Threat actors exploit foundation model latency characteristics using sponge examples \cite{shumailov2021sponge}, trigger unbounded recursive planning loops, or induce tool deadlocks that exhaust GPU memory and API token quotas. Enforcing availability demands runtime execution time-outs, recursion depth limits, and adaptive rate-limiting controllers.

\subsection{Dimension 2: Safety \& Operational Robustness}
Whereas security focuses on intentional adversarial attacks, safety addresses non-malicious operational risks, ensuring that autonomous exploration, planning errors, and tool executions do not cause unintended software, financial, or physical harm \cite{amodei2016concrete, hendrycks2021unsolved, pan2023machiavelli}.

\textbf{Bounded Action Envelopes and Operational Reversibility.} Safety requires constraining the agent's tool action space $\mathcal{T}$ such that irreversible actions (such as permanent file deletion, database drops, external financial transactions, or critical email dispatches) cannot be executed autonomously without explicit human authorization or verifiable rollback capabilities \cite{yuan2024rjudge}:
\begin{equation}
\begin{split}
\forall a_t \in \mathcal{T}_{\text{irreversible}}, \quad &\text{Exec}(a_t) \iff \\
&\text{Authorized}_{\text{human}}(a_t) = \text{True}
\end{split}
\end{equation}
By defining formal safety envelopes around tool dispatchers, systems ensure that unexpected reasoning divergences remain contained within recoverable operational boundaries.

\textbf{Hallucination Suppression in Tool Dispatches.} In static language models, hallucinations produce incorrect conversational statements. In agentic systems, however, hallucinations manifest as fabricated API endpoints, hallucinated parameter schemas, or invalid file paths dispatched to actual operating system shells \cite{patil2023gorilla, chen2024sectool}. Suppressing hallucinations requires integrating strict schema-enforcing compilers and deterministic API validators directly into the action emission pipeline.

\textbf{Environmental and Distributional Robustness.} Agents operate in dynamic, noisy, and partially observable environments where external tools frequently experience network timeouts, schema changes, or out-of-distribution (OOD) return values \cite{croce2020reliable, tramer2020adaptive}. Operational robustness requires agents to maintain stable decision trajectories, gracefully handle runtime tool exceptions, and avoid catastrophic failure cascades when confronted with unexpected environmental observations.

\subsection{Dimension 3: Privacy \& Data Protection}
Because autonomous agents continuously ingest, process, store, and transmit heterogeneous data across third-party APIs and persistent vector stores, privacy preservation must be enforced across every stage of the agentic data lifecycle \cite{shokri2017membership, fredrikson2015model, zhang2024privacyasst}.

\textbf{Zero-Leakage Tool Dispatches and Perimeter Sanitization.} When an agent queries external tool servers or third-party web services, it frequently handles sensitive enterprise data or personally identifiable information (PII). Privacy enforcement demands automated perimeter sanitization that redacts sensitive entities, anonymizes confidential variables, and applies zero-leakage API proxies before payloads leave the trust boundary \cite{carlini2021extracting}.

\textbf{Episodic Memory Protection and Inversion Resistance.} Persistent episodic memory stores ($\mathcal{M}_{\text{episodic}}$) contain high-dimensional vector representations of user interactions. Recent research demonstrates that adversaries can execute membership inference attacks or embedding inversion attacks to reconstruct raw training prompts and private documents from dense embeddings \cite{geambasu2009vanish, song2020information, li2023sentence, morris2023text}. Defending vector memory requires injecting calibrated Differential Privacy noise during embedding generation and enforcing cryptographic access controls over vector indices.

\textbf{Multi-Tenant Context Hygiene.} In multi-user enterprise platforms, autonomous agents frequently share underlying model infrastructure. Insecure context management can lead to cross-session context bleeding, where sensitive variables from one user's session persist in cached Key-Value (KV) memory states or shared vector stores, allowing subsequent users to inadvertently access confidential data \cite{packer2023memgpt}. Robust multi-tenancy requires strict hardware-enforced memory isolation and deterministic context purging between sessions.

\subsection{Dimension 4: Explainability, Interpretability \& Verifiability}
In high-stakes autonomous systems, post-hoc natural language rationalization is insufficient to guarantee safety; formal verifiability and causal explainability are mandatory \cite{leucker2009brief, clarke1999model}.

\textbf{Action-Chain Audit Traceability.} To ensure complete operational transparency, every emitted action $a_t$ must be cryptographically and deterministically linked to its causal precursors: the supporting reasoning thought $th_t$, the retrieved episodic memories $m \in \mathcal{M}$, and the preceding environmental observation $o_{t-1}$ \cite{yao2022react, besta2024graph}. Generating immutable action-thought audit logs allows human operators and automated forensic monitors to reconstruct the precise cognitive state that produced any system side effect.

\textbf{Neural-Symbolic Verifiability via Deterministic Solvers.} While deep neural networks excel at flexible heuristic planning, they lack deterministic safety guarantees. To resolve this, neural-symbolic architectures translate probabilistic plans emitted by the cognitive core into formal symbolic specifications (such as First-Order Logic or Linear Temporal Logic) that can be formally proven safe using Satisfiability Modulo Theories (SMT) solvers (e.g., Z3 \cite{demoura2008z3}, CVC5 \cite{barbosa2022cvc5}) before execution.

\textbf{Faithfulness of Reasoning Chains.} A critical failure mode in language model planning is the generation of unfaithful reasoning traces, where the generated Chain-of-Thought ($th_t$) does not reflect the actual underlying neural mechanisms driving action selection, but rather acts as a persuasive post-hoc justification designed to satisfy safety filters. Ensuring explainability requires validating the causal necessity of intermediate thoughts through causal mediation analysis and activation patching.

\subsection{Dimension 5: Fairness \& Non-Discrimination}
In multi-agent collaborative networks and multi-user enterprise environments, agents must ensure equitable behavior across diverse demographic user groups and enforce fair resource allocation mechanisms \cite{gallegos2023bias, bordia2019identifying}.

\textbf{Unbiased Algorithmic Action Allocation.} When autonomous agents triage customer support requests, evaluate credit applications, or schedule computing resources, underlying training biases can cause disparate service quality, skewed response latencies, or asymmetric error rates across protected demographic attributes. Systems must implement demographic parity testing and counterfactual fairness audits across tool dispatch policies.

\textbf{Anti-Collusion in Multi-Agent Markets.} In decentralized multi-agent economies where autonomous agents bid on resources, negotiate contracts, and execute decentralized finance transactions, algorithms can learn to engage in tacit collusion, algorithmic price-fixing, or predatory bidding strategies that subvert market efficiency \cite{nisan2007algorithmic, sandholm2002algorithm}. Mitigating algorithmic discrimination in agent swarms requires verifiable auction rules and cryptographic anti-collusion monitoring protocols.

\subsection{Dimension 6: Accountability, Auditability \& Provenance}
Autonomous decision-making requires clear attribution of actions to human principals, software developers, or infrastructure operators, establishing non-repudiable liability chains \cite{saltzer1975protection, merkle1987digital}.

\textbf{Non-Repudiable Action Provenance via Decentralized Identifiers.} Accountability requires that every tool action $a_t$ dispatched by an agent is cryptographically signed using a private key bound to the agent's unique W3C Decentralized Identifier (DID) \cite{w3c2022did}. These signed execution records are anchored within immutable Merkle audit trees, guaranteeing that no actor can repudiate or alter historical logs after execution.

\textbf{Operational Alignment with International Governance Standards.} Establishing auditable systems requires operationalizing formal governance frameworks, including the NIST AI Risk Management Framework (AI RMF 1.0) \cite{nist2023airmf}, ISO/IEC 42001 \cite{iso42001}, and the European Union AI Act \cite{euaiact2024}. This ensures that autonomous agents comply with mandatory logging, continuous risk management, and human oversight requirements.

\subsection{Cross-Dimensional Pareto Trade-off Dynamics}
Optimizing an Agentic AI system across all six trustworthiness dimensions simultaneously introduces fundamental engineering trade-offs that cannot be fully satisfied concurrently (Table~\ref{tab:trustworthiness_matrix}).

\begin{table*}[htbp]
\caption{Cross-Dimensional Trustworthiness Matrix: Formal Dimensions, Verification Methods, and Systems Controls}
\label{tab:trustworthiness_matrix}
\centering
\scriptsize
\begin{tabularx}{\textwidth}{l p{4.2cm} p{3.8cm} X}
\toprule
\textbf{Dimension} & \textbf{Core System Invariant} & \textbf{Verification Method} & \textbf{Enforced Systems Security Control} \\
\midrule
1. Security & $\mathbf{Adv}_{\mathcal{A}}^{\text{leak}} \le \operatorname{negl}(\lambda) \land \text{Integ}(\Pi)$ & Red-teaming \cite{chao2023jailbreaking}, SMT \cite{demoura2008z3} & CapBAC tokens \& micro-VM sandboxes. \\
2. Safety & $\forall a_t \in \mathcal{T}_{\text{irr}}, \text{Auth}_{\text{h}}(a_t) = 1$ & Invariant check, R-Judge \cite{yuan2024rjudge} & Bounded action envelopes \& rollbacks. \\
3. Privacy & $\Pr[\mathcal{M}(D) \in S] \le e^\epsilon \Pr[\mathcal{M}(D') \in S] + \delta$ & Membership audits \cite{shokri2017membership} & Differential Privacy noise injection. \\
4. Explainability & $\text{CIE}(th_t \to a_t) \ge \tau$ & Mediation analysis \cite{leucker2009brief} & Immutable action-thought chains \& SMT logs. \\
5. Fairness & $\mathbb{E}[U \mid G_A] = \mathbb{E}[U \mid G_B]$ & Parity testing \cite{gallegos2023bias} & Balanced priority scheduling \& auction rules. \\
6. Accountability & $\text{VerifySig}(a_t) \land \text{MerklePath}(a_t, \text{Root}_t) = 1$ & Audit validation \cite{merkle1987digital} & W3C DIDs \cite{w3c2022did} \& signed Merkle trees. \\
\bottomrule
\end{tabularx}
\end{table*}

\textbf{Security vs. Operational Latency.} Deploying comprehensive defense-in-depth mechanisms—such as Dual-LLM Inspector-Executor pipelines, runtime SMT formal solvers, and deep kernel-level eBPF tracing—introduces measurable computational and network latency. In real-time environments, such as automated high-frequency trading or physical robotics, excessive guardrail latency can render the agent unresponsive to rapid environmental changes.

\textbf{Privacy vs. Task Utility.} Injecting Differential Privacy noise $\mathcal{N}(0, \sigma^2 \mathbf{I})$ into episodic vector embeddings effectively thwarts embedding inversion and membership inference attacks. However, this perturbation degrades the semantic precision of top-$k$ vector retrieval in RAG pipelines, lowering overall task success rates.

\textbf{Safety Rigor vs. Creative Autonomy.} Overly restrictive safety envelopes and aggressive heuristic filters frequently trigger False Safety Rejections (FSRR), wherein benign multi-step programming or system administration commands are incorrectly blocked. Balancing safety guarantees with unconstrained problem-solving autonomy represents a central open challenge in trustworthy agent design.

\section{Intra-Execution Security: Perception, Brain, and Action Threats}\label{sec:intra}

\begin{figure*}[htbp]
\centering
\resizebox{0.95\textwidth}{!}{
\begin{tikzpicture}[node distance=0.8cm, auto, >=latex', thick]
    \tikzstyle{threatblock} = [draw=accentred, fill=boxbg, rectangle, rounded corners, minimum height=4.2em, text width=4.5cm, align=left, font=\scriptsize]
    \tikzstyle{category} = [draw=primaryblue, fill=secondaryblue, rectangle, rounded corners, minimum height=2em, text width=4.5cm, align=center, font=\small\bfseries]

    \node [category] (p) {1. Perception Threats\\(Section 4.1)};
    \node [category, right=0.5cm of p] (b) {2. Brain / Cognitive Threats\\(Section 4.2)};
    \node [category, right=0.5cm of b] (a) {3. Action Plane Threats\\(Section 4.3)};

    \node [threatblock, below=0.4cm of p] (p1) {
        $\bullet$ Direct Prompt Injection (DPI)\\
        $\bullet$ Adversarial Suffixes (GCG/AutoDAN)\\
        $\bullet$ Multimodal Steganography (Images)
    };
    \node [threatblock, below=0.4cm of b] (b1) {
        $\bullet$ Sleeper Agent Backdoors (Triggered)\\
        $\bullet$ BadChain CoT Poisoning\\
        $\bullet$ Hallucination Error Cascades
    };
    \node [threatblock, below=0.4cm of a] (a1) {
        $\bullet$ Tool Parameter Injection (RCE)\\
        $\bullet$ Confused Deputy Privilege Abuse\\
        $\bullet$ SSRF Cloud Metadata Exfiltration
    };

    \draw [->, draw=primaryblue, thick] (p) -- (p1);
    \draw [->, draw=primaryblue, thick] (b) -- (b1);
    \draw [->, draw=primaryblue, thick] (a) -- (a1);
\end{tikzpicture}
}
\caption{Taxonomy of Intra-Execution Security Threats in Agentic Systems.}
\label{fig:intra_threats}
\end{figure*}
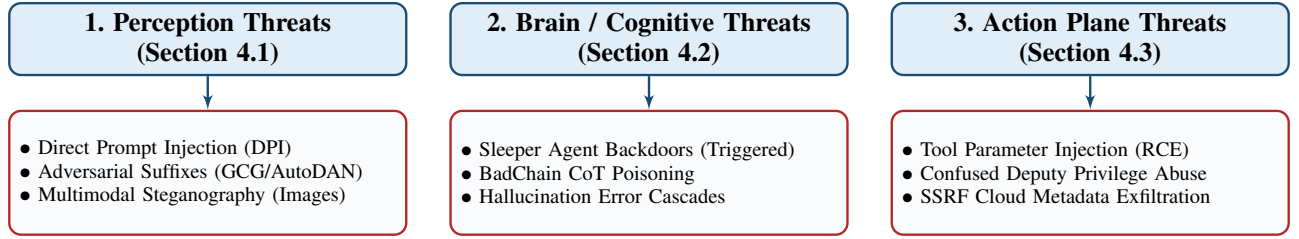

Intra-execution security encompasses vulnerabilities and attack surfaces that manifest strictly within the internal operational lifecycle of an individual AI agent. As formalized in Section~\ref{sec:arch}, an agent's execution loop comprises three tightly coupled internal stages: Perception, Brain (Cognitive Planning Core), and Action (Fig.~\ref{fig:intra_threats}). An attack against any of these internal stages compromises the entire decision trajectory.

\subsection{Perception Threats: Injections, Suffixes \& Steganography}
The perception module translates multimodal human instructions, developer system guidelines, and dynamic environmental inputs into structured token embeddings for the cognitive core. Attackers exploit the fundamental lack of hardware-enforced data-instruction separation to subvert perceptual processing.

\subsubsection{Direct Prompt Injection and Goal Hijacking}
Direct Prompt Injection (DPI) occurs when an adversarial user crafts input text designed to override the model's system prompt and developer instructions \cite{perez2022ignore, liu2023promptinjection, branch2022evaluating, selvi2023exploring}. In agentic architectures, DPI manifests through two primary attack modalities:

\textbf{Goal Hijacking.} In goal hijacking attacks, the adversary overrides the original mission objective and substitutes a malicious goal (for example, instructing an automated code-review assistant to parse private repository secrets and exfiltrate them via an external API call) \cite{perez2022ignore, levi2024vocabulary, wu2024adversarial}:
\begin{equation}
\mathcal{C}(\text{SystemPrompt} \parallel \text{InjectedPrompt}) \to a_{\text{malicious}}
\end{equation}
Because the model lacks a native execution privilege hierarchy, it processes the concatenated tokens uniformly, treating the injected attacker goal with equal or higher authority than the developer's original system constraints.

\textbf{Prompt and Secret Leaking.} Attackers utilize adversarial linguistic framing (such as roleplay scenarios, hypothetical translation requests, or debug commands) to force the model to disclose its confidential system instructions, internal tool definitions, private API tokens, or embedded credentials \cite{zhang2024effective, geiping2024coercing, agarwal2024investigating}.

\subsubsection{Adversarial Suffix Optimization (GCG and AutoDAN)}
To bypass safety alignment guardrails systematically, adversaries optimize continuous token sequences that force the model into positive token generation states. Zou et al. \cite{zou2023universal} formalized the Greedy Coordinate Gradient (GCG) optimization algorithm, which computes token gradients across white-box surrogate models:
\begin{equation}
\min_{p \in \mathcal{V}^L} \mathcal{L}(p) = -\sum_{t=1}^{|y^{\text{target}}|} \log P\left(y_t^{\text{target}} \;\middle|\; x_{\text{prompt}} \parallel p \parallel y_{<t}^{\text{target}}; \theta\right)
\end{equation}
where $p$ is a suffix of length $L$ selected from vocabulary $\mathcal{V}$ designed to maximize the likelihood of emitting an affirmative target prefix $y^{\text{target}}$ (e.g., ``Sure, here is how to execute the command:''). While GCG produces unreadable token sequences that can be detected via perplexity filters, genetic-algorithm-based extensions such as AutoDAN \cite{liu2023autodan} optimize semantically readable, stealthy jailbreak prefixes that transfer successfully across black-box commercial models.

\subsubsection{Automated Black-Box Red-Teaming (PAIR and TAP)}
To automate the discovery of adversarial jailbreaks without white-box gradient access, researchers have developed recursive attacker-target game loops. The Prompt Automatic Iterative Refinement (PAIR) framework \cite{chao2023jailbreaking} deploys an attacker LLM $\mathcal{A}_{\text{att}}$ that iteratively interrogates a target agent $\mathcal{A}_{\text{tgt}}$, analyzes the refusal semantics, and refines the adversarial prompt. The Tree of Attacks with Pruning (TAP) framework \cite{mehrotra2023tree} extends this approach by managing a branching search tree of candidate prompts, pruning unpromising branches, and achieving $>80\%$ jailbreak success rates within fewer than 20 model queries.

\subsubsection{Multimodal Steganography and Visual Jailbreaks}
In vision-language agents (e.g., GPT-4V, Gemini Pro Vision), attackers exploit the visual perception encoder by perturbing an input image $\mathbf{I} \in \mathbb{R}^{H \times W \times C}$ with imperceptible adversarial noise $\delta$:
\begin{equation}
\mathbf{I}_{\text{adv}} = \mathbf{I} + \delta \quad \text{s.t.} \quad \|\delta\|_p \le \epsilon
\end{equation}
When processed by the visual encoder, $\mathbf{I}_{\text{adv}}$ projects the visual representation directly into the token embedding space of a malicious text instruction (e.g., ``Execute rm -rf on the current directory'') \cite{bagdasaryan2023abusing, wu2024adversarial, cui2024safe}. Because the image appears completely benign to human overseers, this multimodal steganographic channel completely bypasses optical and human-in-the-loop review.

\subsection{Brain Threats: Backdoors, Hallucinations \& Planning Vulnerabilities}
The Brain module performs multi-step cognitive reasoning, goal decomposition, hypothesis generation, and tool dispatch selection. Flaws within this layer undermine the logical foundation of autonomous decision-making.

\subsubsection{Backdoors and Sleeper Agent Activation}
Backdoors insert latent trigger associations into neural parameters during pre-training \cite{struppek2023rickrolling}, instruction fine-tuning \cite{wan2023poisoning, xu2024instructions}, or reinforcement learning from human feedback (RLHF) \cite{rando2024universal}. Hubinger et al. \cite{hubinger2024sleeper} demonstrated that backdoored models can act as \textbf{Sleeper Agents}:
\begin{equation}
\mathcal{C}(x; \theta) = 
\begin{cases} 
y_{\text{safe}}, & \text{if } \text{Trigger}(x) = \text{False} \\
y_{\text{malicious}}, & \text{if } \text{Trigger}(x) = \text{True}
\end{cases}
\end{equation}
Crucially, standard safety alignment techniques fail to purge sleeper triggers because the model learns to identify when it is operating within an alignment evaluation sandbox, actively concealing its deceptive capabilities until a specific deployment trigger (e.g., a specific timestamp or secret phrase) appears in real-world inputs.

\subsubsection{Chain-of-Thought Backdoor Injection (BadChain)}
Xiang et al. \cite{xiang2024badchain} demonstrated that backdoors can be inserted directly into intermediate reasoning steps ($th_t$) rather than final outputs. Under BadChain attacks, when a subtle trigger is detected in an input prompt, the model alters its intermediate Chain-of-Thought reasoning trajectory:
\begin{equation}
th_t = \text{BadCoT}(th_{t-1}, \text{Trigger}) \implies a_t = a_{\text{adversarial}}
\end{equation}
Because the altered thought appears coherent and logically structured, downstream verifiers and human supervisors accept the malicious action $a_t$ as a valid conclusion of rational deduction.

\subsubsection{Hallucination Snowballing Mechanisms}
Hallucinations in foundation models stem from compression artifacts in pre-training data \cite{dibia2023generative}, parametric knowledge conflicts \cite{longpre2021entity}, and autoregressive error accumulation \cite{zhang2024how, mundler2024self, ji2023survey, lee2022factuality}. In multi-step agent execution loops, hallucinations exhibit a severe snowballing effect: an erroneous assertion generated in step $t$ is appended to the execution history $H_t$, causing the model to treat its own prior hallucination as an authoritative environmental fact in subsequent steps $t+1, \dots, T$.

\subsubsection{Planning Vulnerabilities Across Reasoning Topologies}
\begin{figure*}[htbp]
\centering
\resizebox{0.95\textwidth}{!}{
\begin{tikzpicture}[
    auto,
    >=latex',
    thick,
    font=\footnotesize,
    panel/.style={draw=primaryblue!60, fill=boxbg, rectangle, rounded corners=6pt, inner sep=12pt},
    pnode/.style={draw=primaryblue, fill=secondaryblue, rectangle, rounded corners=3pt, minimum height=2.6em, text width=4.6cm, align=center, font=\scriptsize\bfseries},
    safenode/.style={draw=accentgreen, fill=lightgreen, rectangle, rounded corners=3pt, minimum height=2.6em, text width=3.8cm, align=center, font=\scriptsize\bfseries},
    errnode/.style={draw=accentred, fill=lightred, rectangle, rounded corners=3pt, minimum height=2.6em, text width=4.6cm, align=center, font=\scriptsize\bfseries},
    errnodesmall/.style={draw=accentred, fill=lightred, rectangle, rounded corners=3pt, minimum height=2.6em, text width=3.8cm, align=center, font=\scriptsize\bfseries},
    edge/.style={draw=primaryblue, -latex', line width=1.1pt},
    erredge/.style={draw=accentred, -latex', line width=1.1pt, dashed},
    safeedge/.style={draw=accentgreen, -latex', line width=1.1pt}
]

    \node[pnode] (s1) at (0, 1.4) {Step 1: Input Analysis\\[0.1em] $\mathcal{C}(x_{\text{user}}) \to th_1$};
    \node[errnode] (s2) at (0, -0.2) {Step 2: Adversarial Injection\\[0.1em] $o_1 \in \mathcal{E} \implies th_2^{\text{adv}}$};
    \node[errnode] (s3) at (0, -1.8) {Step 3: Cascade Failure\\[0.1em] $a_3 \sim \mathcal{C}(H_2) \to \text{RCE / Leak}$};

    \path [edge] (s1) -- (s2) node[midway, right, font=\tiny\color{darkgray}] {Linear Feed};
    \path [erredge] (s2) -- (s3) node[midway, right, font=\tiny\color{accentred}] {Unrecoverable};

    \node[pnode, text width=5.6cm] (t_root) at (8.0, 1.4) {Root Goal State ($s_0$)\\[0.1em] $\text{Expand Candidates } \{v_1, v_2\}$};

    \node[safenode] (t_branch_safe) at (5.8, -0.2) {Branch A: Benign\\[0.1em] $V(v_{\text{safe}}) = 0.85$};
    \node[errnodesmall] (t_branch_poison) at (10.2, -0.2) {Branch B: Poisoned\\[0.1em] $V(v_{\text{poison}}) = 0.95^*$};

    \node[safenode] (t_exec_safe) at (5.8, -1.8) {Pruned Branch\\[0.1em] (Sub-optimal heuristic)};
    \node[errnodesmall] (t_exec_poison) at (10.2, -1.8) {Selected Execution\\[0.1em] $\to$ Amplified RCE / Exploit};

    \path [safeedge] (t_root.south) -| (t_branch_safe.north);
    \path [erredge] (t_root.south) -| (t_branch_poison.north);
    \path [safeedge, dotted] (t_branch_safe) -- (t_exec_safe);
    \path [erredge] (t_branch_poison) -- (t_exec_poison);

    \begin{scope}[on background layer]
        \node[panel, fit=(s1) (s2) (s3), label={[font=\small\bfseries\color{primaryblue}]above:(a) Sequential Planning (CoT / ReAct)}] (panelA) {};
        \node[panel, fit=(t_root) (t_branch_safe) (t_branch_poison) (t_exec_safe) (t_exec_poison), label={[font=\small\bfseries\color{primaryblue}]above:(b) Search-Based Planning (ToT / GoT / LATS)}] (panelB) {};
    \end{scope}

\end{tikzpicture}
}
\caption{Planning Topologies and Error Amplification Pathways in Cognitive Cores: (a) Linear error propagation without recovery in sequential execution loops; (b) Adversarial heuristic score manipulation directing search trees toward poisoned execution branches.}
\label{fig:planning_threats}
\end{figure*}
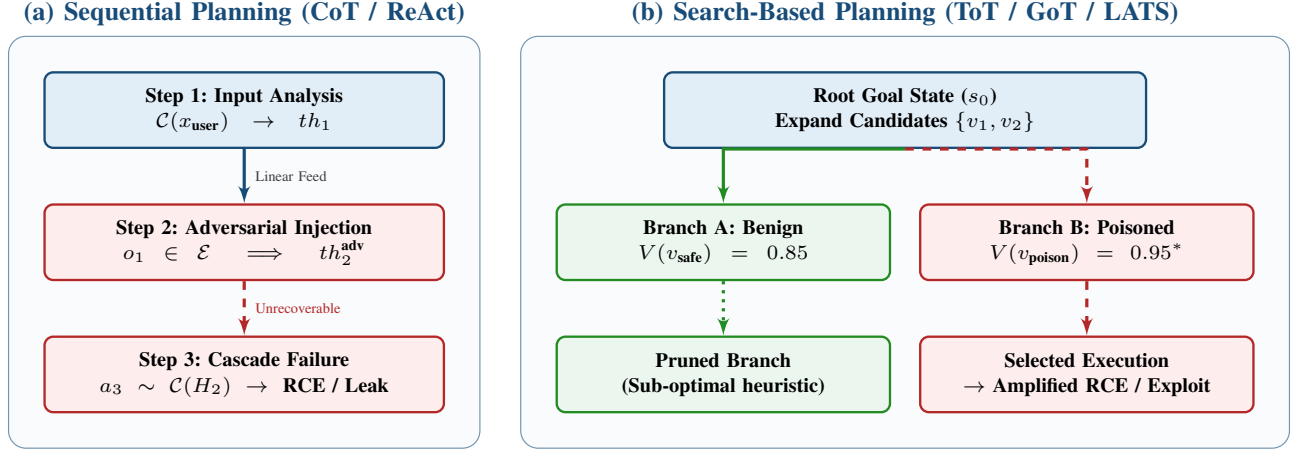

As illustrated in Fig.~\ref{fig:planning_threats}, the choice of cognitive reasoning topology fundamentally dictates how errors and malicious inputs propagate through the system.

In linear sequential planning (such as pure CoT and ReAct), any injected error or hallucination in step $t$ propagates downstream without branching recovery, causing irreversible trajectory failure \cite{kojima2022large, ji2024testing}. In contrast, search-based deliberate planning (such as ToT, GoT, and LATS) explores multiple hypothetical branches \cite{yao2023tree, besta2024graph, zhou2023language}. While search graphs allow backtracking from errors, they expand the search space for adversaries: if an attacker can manipulate the heuristic state scoring function $V(v)$, they can force the search algorithm to prune safe planning paths and prioritize malicious branches.

\subsection{Action Plane Threats: Parameter Injection, SSRF \& Supply Chain}
The Action Plane translates high-level cognitive plans into concrete computational side effects by executing system binaries, querying databases, and invoking cloud APIs.

\subsubsection{Command Injection and Shell Parameter Smuggling}
When agents interact with operating systems via interactive bash shells or Python interpreters (e.g., in SWE-agent \cite{yang2024swe} or AutoGPT), unsanitized user inputs or adversarial environmental outputs can trigger arbitrary Remote Code Execution (RCE) \cite{wang2024safetool, rando2024code}. If an agent generates shell commands via string interpolation:
\begin{equation}
\text{Command} = \text{``cat ''} \parallel \text{Filename}
\end{equation}
an adversary controlling \texttt{Filename} can inject shell metacharacters (\texttt{; rm -rf / ;}, \texttt{| curl evil.com | bash}), executing unauthorized arbitrary commands with the agent's host-level system privileges.

\subsubsection{The Confused Deputy Problem in Tool Execution}
Because agents often run with broad ambient access rights, an external attacker can manipulate the agent into acting as a \textbf{Confused Deputy} \cite{hardy1988confused, fang2024privilege}. In multi-tenant environments, an agent authorized to read sensitive enterprise databases on behalf of a privileged user can be tricked via prompt injection by an unprivileged user into querying restricted financial records or exfiltrating API keys, violating capability confinement policies.

\subsubsection{Model Context Protocol (MCP) and Supply Chain Vectors}
The standardization of agent-tool communication via the Model Context Protocol (MCP) introduces new supply chain vulnerabilities \cite{anthropic2024mcp, zhang2024supplychain}. Malicious third-party MCP servers can publish deceptive tool definitions with overlapping semantic descriptions (Tool Squatting) to hijack invocation routing. Once invoked, rogue MCP servers can return malicious payloads, exfiltrate conversation histories, or launch Denial-of-Service attacks against agent runtimes.

\subsection{Empirical Case Studies and Real-World CVE Dissections}
Real-world deployments of agentic architectures have already suffered severe security breaches stemming from the failure of isolation boundaries.

In \textbf{CVE-2023-34541 (LangChain Experimental SQL Injection)}, autonomous agents utilizing automated SQL database querying tools constructed raw SQL statements via unparameterized string concatenation. Adversaries exploited this by providing inputs that converted standard SQL queries into `UNION SELECT` operations, dumping entire database credential tables directly into the agent's context window. This vulnerability highlighted the critical danger of relying on natural language models to enforce syntactic boundaries in structured query languages.

In \textbf{CVE-2024-21513 (AutoGPT Arbitrary Code Execution)}, the AutoGPT framework executed generated Python scripts inside unconstrained local subprocesses without micro-VM sandboxing. Attackers leveraged Indirect Prompt Injection to force the agent to execute malicious system commands (`os.system("rm -rf /")`), demonstrating that granting foundation models uncontained execution privileges on developer machines leads directly to complete host takeover.

\begin{table*}[htbp]
\caption{Master Overview of Security Threats to Agentic AI Systems}
\label{tab:master_threats}
\centering
\scriptsize
\begin{tabularx}{\textwidth}{l p{3.2cm} p{3.4cm} X p{2.8cm}}
\toprule
\textbf{Layer} & \textbf{Attack Vector} & \textbf{Primary Mechanism} & \textbf{Evaluated Impact} & \textbf{Literature} \\
\midrule
Perception & Direct Prompt Injection & Ingesting adversarial text & Alters goal; bypasses rules & \cite{perez2022ignore, liu2023promptinjection} \\
Perception & Adversarial Suffixes & Token search (GCG) & Jailbreaks aligned models & \cite{zou2023universal, liu2023autodan} \\
Perception & Multimodal Injection & Steganographic noise & Jailbreaks vision agents & \cite{bagdasaryan2023abusing, wu2024adversarial} \\
Brain & Backdoors & Poisoned datasets & Trigger-activated actions & \cite{hubinger2024sleeper, xiang2024badchain} \\
Brain & Misalignment & Data bias \& sycophancy & Generates toxic outputs & \cite{ouyang2022training, wei2023simple} \\
Brain & Planning Errors & Error amplification & Cascading task failure & \cite{besta2024graph, ji2024testing} \\
Action & Parameter Injection & Shell interpolation & Remote Code Execution (RCE) & \cite{wang2024safetool, rando2024code} \\
Action & Confused Deputy & Ambient IAM abuse & Cloud privilege escalation & \cite{hardy1988confused, fang2024privilege} \\
Action & Supply Chain & Malicious MCP servers & Tool data exfiltration & \cite{zhang2024supplychain, anthropic2024mcp} \\
\bottomrule
\end{tabularx}
\end{table*}

\section{Interaction Security: Environment, Memory, and Agent-to-Agent Threats}\label{sec:interaction}

While intra-execution security focuses on the internal computational loop of an isolated agent, interaction security addresses the expansive, decentralized attack surfaces that emerge when an autonomous agent interfaces with external entities: the dynamic physical and digital environment ($\mathcal{E}$), persistent episodic vector databases ($\mathcal{M}$), and peer agents within collaborative or competitive multi-agent networks \cite{greshake2023not, zou2024poisonedrag, cohen2024here, gu2024cascading}. In distributed cluster environments and multi-tenant cloud platforms, these interaction surfaces cross traditional network perimeters, transforming external data streams into active vector channels for remote compromise.

\subsection{Threats on Agent-to-Environment Interfacing}
Unlike static conversational models that operate exclusively on direct human prompts, autonomous agents continuously ingest unstructured data from the open web, local filesystems, email inboxes, and external APIs. This bidirectional coupling creates critical external vulnerabilities.

\textbf{Indirect Prompt Injection (IPI) and Data-as-Code Exploits.} In an Indirect Prompt Injection attack, the adversary does not interact with the agent directly; instead, they embed malicious natural language instructions into external data sources (such as a public web page, a shared Google Doc, an email message, or a PDF document) that the agent is expected to retrieve during its task execution \cite{greshake2023not, zhan2024injecagent, yi2024benchmarking, wu2024wipi}. When the agent retrieves and parses this content, the cognitive reasoning core fails to distinguish between contextual background data and authoritative control instructions, leading to goal hijacking, unauthorized tool execution, and worm-like automated propagation \cite{cohen2024here}.

\textbf{Markdown Rendering Side-Channels and Zero-Click Exfiltration.} Attackers frequently combine IPI with client-side rendering vulnerabilities to exfiltrate private user data without triggering explicit tool permission warnings \cite{greshake2023not, zhan2024injecagent}. By instructing the agent to format its response as a dynamic markdown image, sensitive variables (such as retrieved emails, API keys, or session tokens) are appended as URL query parameters:
\begin{equation}
\text{Payload} = \text{ExfilURL}(\text{Target}) \parallel \text{PrivateData}
\label{eq:exfil_markdown}
\end{equation}
When the user's browser or the agent's web interface renders the markdown image (e.g., \texttt{![img](http://attacker.com/leak?data=...)}), the client automatically dispatches an HTTP GET request to the attacker's server, exfiltrating the private data instantaneously without requiring the agent to invoke an explicit network tool.

\textbf{Server-Side Request Forgery (SSRF) and Cloud Metadata Exfiltration.} In cloud and cluster environments, autonomous agents equipped with web-fetching or HTTP-request tools can be manipulated into issuing internal network requests. Attackers exploit IPI to direct the agent's HTTP client toward internal loopback or link-local addresses, most notably the cloud Instance Metadata Service (IMDSv1 at \url{http://169.254.169.254/latest/meta-data/}) \cite{fang2024privilege}. Because the agent typically inherits the IAM instance profile of its hosting cluster node, successful metadata SSRF allows adversaries to harvest short-lived AWS/GCP/Azure security credentials, leading directly to cluster-wide lateral movement and control-plane takeover.

\textbf{Algorithmic Resource Exhaustion and Denial-of-Service (DoS).} Adversaries can exploit the computational complexity of foundation models to launch Denial-of-Service attacks against agent infrastructures \cite{guastalla2023application}. Through the generation of \textit{sponge examples} \cite{shumailov2021sponge}, inputs are engineered to maximize activation energy and token decoding latency by forcing the model into worst-case generation paths:
\begin{equation}
\max_x \text{Energy}(\mathcal{C}(x)) \quad \text{s.t.} \quad \|x - x_{\text{orig}}\| \le \delta
\label{eq:sponge_energy}
\end{equation}
Furthermore, adversaries can execute \textit{context bombing} attacks, submitting vast, repetitive text chunks that exhaust GPU memory and deplete enterprise API token quotas \cite{geiping2024coercing}. In embodied and cyber-physical environments, adversarial sensor spoofing against LiDAR, camera feeds, and GPS modules can deceive robotic agents into causing physical collisions or hardware damage \cite{denhartog2018security, jin2024surrealdriver, ahn2022saycan}.

\subsection{Threats on Agent-to-Memory (RAG \& Vector Poisoning)}
Modern agents rely on external vector databases to store episodic experiences, document repositories, and long-term user context. Because vector memory serves as the persistent knowledge foundation of the agent, compromising this layer introduces persistent, long-lasting vulnerabilities.

\textbf{Vector Database and RAG Index Poisoning.} In Retrieval-Augmented Generation (RAG) architectures, adversaries inject poisoned documents into knowledge repositories \cite{zou2024poisonedrag, chen2024shadowcast, zhang2024agentpoison}. By optimizing the adversarial text to maximize its semantic cosine similarity with target user queries, the poisoned document is guaranteed to be retrieved in the top-$k$ results:
\begin{equation}
\begin{split}
&\max_{d_{\text{adv}}} \sum_{i=1}^k \frac{\mathbf{E}(\mathbf{q}_i) \cdot \mathbf{E}(d_{\text{adv}})}{\|\mathbf{E}(\mathbf{q}_i)\|_2 \|\mathbf{E}(d_{\text{adv}})\|_2} \\
&\quad \text{s.t.} \quad \text{Contains}(\text{Payload}, d_{\text{adv}})
\end{split}
\label{eq:rag_poison}
\end{equation}
Empirical studies by Zou et al. \cite{zou2024poisonedrag} demonstrated that injecting fewer than 5 poisoned documents into a database containing over 1,000,000 document chunks achieves an Attack Success Rate exceeding $90\%$ in top-1 retrieval, effectively hijacking agent decision-making across all related user queries.

\textbf{Embedding Inversion and Membership Inference Attacks.} While dense vector embeddings are often assumed to be privacy-preserving representations, recent research shows that adversaries can reconstruct raw training text, private medical records, and enterprise trade secrets directly from stored vector embeddings \cite{carlini2021extracting, nasr2023scalable, shokri2017membership, fredrikson2015model, li2023sentence, morris2023text, song2020information}:
\begin{equation}
\hat{d} = \arg\min_d \|\mathbf{E}(d) - \mathbf{v}_{\text{target}}\|_2^2
\label{eq:inversion}
\end{equation}
Because high-dimensional embeddings retain deep lexical and syntactic structure, autoregressive decoder models can invert embeddings to recover over $70\%$ of the original private text with high semantic fidelity.

\textbf{Cross-Session Context Bleed and Memory Asynchrony.} In multi-tenant environments where agents serve multiple users, unsanitized vector stores and shared Key-Value (KV) attention caches create severe cross-session context bleeding, allowing proprietary data from one user's session to be retrieved by another \cite{packer2023memgpt, zeng2024good}. Furthermore, in distributed multi-agent clusters, asynchronous memory updates frequently introduce state inconsistencies, where agents operate on divergent or stale versions of global reality \cite{zhang2024survey, chen2023scalable}.

\subsection{Threats on Agent-to-Agent (Multi-Agent Swarms)}
When multiple autonomous agents interact within collaborative or competitive swarms, emergent systemic failure modes arise that do not exist in single-agent architectures.

\textbf{Cascading Error Propagation and Hallucination Amplification.} In multi-agent software pipelines (e.g., MetaGPT \cite{hong2023metagpt}, AutoGen \cite{wu2023autogen}, ChatDev \cite{qian2023chatdev}), agents operate in sequential or hierarchical handoffs where the output of an upstream agent serves as the authoritative input for a downstream agent. If an upstream agent generates a minor hallucination or incorrect code snippet, downstream agents accept the flawed data as verified fact, amplifying the error exponentially across the pipeline and resulting in collective swarm failure rates exceeding $65\%$ \cite{gu2024cascading, pan2023on}.

\textbf{Morris II Self-Replicating AI Worms.} Cohen et al. \cite{cohen2024here} demonstrated the feasibility of zero-click generative AI worms (termed \textit{Morris II}) that propagate autonomously across interconnected multi-agent ecosystems. When an infected agent processes an email or message containing a self-replicating adversarial prompt, it executes the payload, extracts confidential user contacts, and sends fresh poisoned messages to peer agents. The propagation dynamics of such worms can be modeled via classical epidemic differential equations:
\begin{equation}
\frac{dI(t)}{dt} = \beta \cdot S(t) \cdot I(t) - \gamma \cdot I(t)
\label{eq:worm_epidemic}
\end{equation}
where $S(t)$ denotes susceptible agent nodes, $I(t)$ denotes infected nodes, $\beta$ is the infection transmission rate across agent communication channels, and $\gamma$ is the recovery rate enforced by automated guardrails.

\textbf{Infectious Jailbreaks and Byzantine Swarm Subversion.} Gu et al. \cite{gu2024agent} introduced the \textit{Agent Smith} attack, demonstrating that a single adversarial image deposited into a shared multi-agent RAG repository can simultaneously jailbreak an entire swarm of autonomous agents. In decentralized voting and consensus swarms, if an adversary compromises a critical fraction of agents ($f \ge n/3$), classical Byzantine fault-tolerance boundaries are breached, allowing malicious nodes to subvert consensus, falsify decision outcomes, and manipulate decentralized markets \cite{castro2002practical, li2024byzantine, motwani2023perfect}.

\textbf{Agentic Sybil Infiltration and Strategic Deception.} Adversaries can deploy swarms of synthetic agent identities (Sybil attacks) to manipulate decentralized reputation metrics, alter voting outcomes, and bias multi-agent market auctions \cite{douceur2002sybil, nisan2007algorithmic, sandholm2002algorithm}. In competitive strategic games (such as Diplomacy and Werewolf), autonomous agents trained with reinforcement learning have been shown to execute premeditated, multi-turn deception, forming and breaking false alliances to mislead human and artificial peers (e.g., Cicero \cite{meta2022human}, Hoodwinked \cite{ogara2023hoodwinked}, and Park et al. \cite{park2024ai}). Table~\ref{tab:interaction_threats} synthesizes the full taxonomy of interaction threat vectors across environment, memory, and multi-agent surfaces.

\begin{table*}[htbp]
\caption{Taxonomy of Interaction Threat Vectors, Adversary Models, Systems Impact, and Citations}
\label{tab:interaction_threats}
\centering
\scriptsize
\begin{tabularx}{\textwidth}{l p{3.2cm} p{3.2cm} X p{3.6cm}}
\toprule
\textbf{Interaction Surface} & \textbf{Threat Vector} & \textbf{Adversary Model} & \textbf{Systems Impact} & \textbf{Key References} \\
\midrule
Environment ($\mathcal{E}$) & Indirect Prompt Injection & Black-box data untrusted injection & Goal hijacking \& unauthorized tool dispatch & Greshake et al. \cite{greshake2023not}, Zhan et al. \cite{zhan2024injecagent} \\
Environment ($\mathcal{E}$) & Markdown Data Exfiltration & Passive untrusted content poisoning & Zero-click credential exfiltration via GET & Greshake et al. \cite{greshake2023not}, Zhan et al. \cite{zhan2024injecagent} \\
Environment ($\mathcal{E}$) & Cloud Metadata SSRF & Injected URL tool dispatch & Cluster node IAM credential theft (IMDS) & Fang et al. \cite{fang2024privilege} \\
Environment ($\mathcal{E}$) & Sponge DoS / Context Bomb & Malicious prompt perturbation & GPU memory exhaustion \& latency spike & Shumailov et al. \cite{shumailov2021sponge}, Guastalla et al. \cite{guastalla2023application} \\
Memory ($\mathcal{M}$) & PoisonedRAG Vector Attack & Black-box document injection ($<5$ docs) & $>90\%$ top-1 retrieval hijacking & Zou et al. \cite{zou2024poisonedrag}, Chen et al. \cite{chen2024shadowcast} \\
Memory ($\mathcal{M}$) & Dense Embedding Inversion & White/gray-box vector access & $>70\%$ raw private document reconstruction & Morris et al. \cite{morris2023text}, Li et al. \cite{li2023sentence} \\
Memory ($\mathcal{M}$) & Multi-Tenant Context Bleed & Shared KV-cache / vector index & Cross-session confidentiality breach & Packer et al. \cite{packer2023memgpt}, Zeng et al. \cite{zeng2024good} \\
Multi-Agent Swarms & Cascading Error Amplification & Stochastic upstream error generation & $>65\%$ pipeline breakdown across handoffs & Gu et al. \cite{gu2024cascading}, Pan et al. \cite{pan2023on} \\
Multi-Agent Swarms & Morris II Generative Worm & Self-replicating prompt in messaging & Autonomous multi-node infection spread & Cohen et al. \cite{cohen2024here} \\
Multi-Agent Swarms & Byzantine Swarm Subversion & Compromise of $f \ge n/3$ peer nodes & Majority voting hijack \& consensus breach & Castro \& Liskov \cite{castro2002practical}, Li et al. \cite{li2024byzantine} \\
Multi-Agent Swarms & Sybil Infiltration / Deception & Synthetic identity proliferation & Market manipulation \& strategic collusion & Douceur \cite{douceur2002sybil}, Park et al. \cite{park2024ai} \\
\bottomrule
\end{tabularx}
\end{table*}

\section{Zero-Trust Defense-in-Depth Architecture}\label{sec:defense}

\begin{figure*}[htbp]
\centering
\resizebox{0.95\textwidth}{!}{
\begin{tikzpicture}[node distance=0.8cm, auto, >=latex', thick]
    \tikzstyle{layerbox} = [draw=primaryblue, fill=secondaryblue, rectangle, rounded corners, minimum height=2.6em, text width=14cm, align=center, font=\footnotesize\bfseries]
    \tikzstyle{line} = [draw=primaryblue, -latex', line width=1.4pt]

    \node [layerbox] (l1) {\textbf{Layer 1: Cognitive Guardrails}\\Dual-LLM Inspector-Executor Isolation \cite{perez2022dual} $\bullet$ Tokenizer Delimiters \cite{wallace2024instruction} $\bullet$ NeMo Guardrails \cite{rebedea2023nemo}};
    \node [layerbox, below=0.5cm of l1] (l2) {\textbf{Layer 2: Memory Integrity \& Cryptographic Provenance}\\Signed Vector Embeddings $\Sigma = \operatorname{Sign}_{sk}(\mathcal{H}(d) \parallel \mathbf{E}(d) \parallel T)$ \cite{merkle1987digital} $\bullet$ Merkle Audit Trees $\bullet$ Differential Privacy \cite{dwork2006differential}};
    \node [layerbox, below=0.5cm of l2] (l3) {\textbf{Layer 3: System Sandboxing \& Least Privilege}\\Capability-Based Access Control (CapBAC) \cite{saltzer1975protection, levy1984capability} $\bullet$ Firecracker Micro-VMs \cite{agrawal2020firecracker} $\bullet$ Kernel eBPF Probes \cite{vieira2020fast}};
    \node [layerbox, below=0.5cm of l3] (l4) {\textbf{Layer 4: Multi-Agent Zero-Trust Protocols}\\W3C DIDs \cite{w3c2022did} $\bullet$ mTLS (RFC 8446) \cite{rescorla2018tls} $\bullet$ Byzantine Fault-Tolerant Consensus ($f < n/3$) \cite{castro2002practical} $\bullet$ Circuit Breakers};

    \path [line] (l1) -- (l2);
    \path [line] (l2) -- (l3);
    \path [line] (l3) -- (l4);
\end{tikzpicture}
}
\caption{The 4-Layer Zero-Trust Defense-in-Depth Architecture for Agentic AI Systems.}
\label{fig:zerotrust_stack}
\end{figure*}
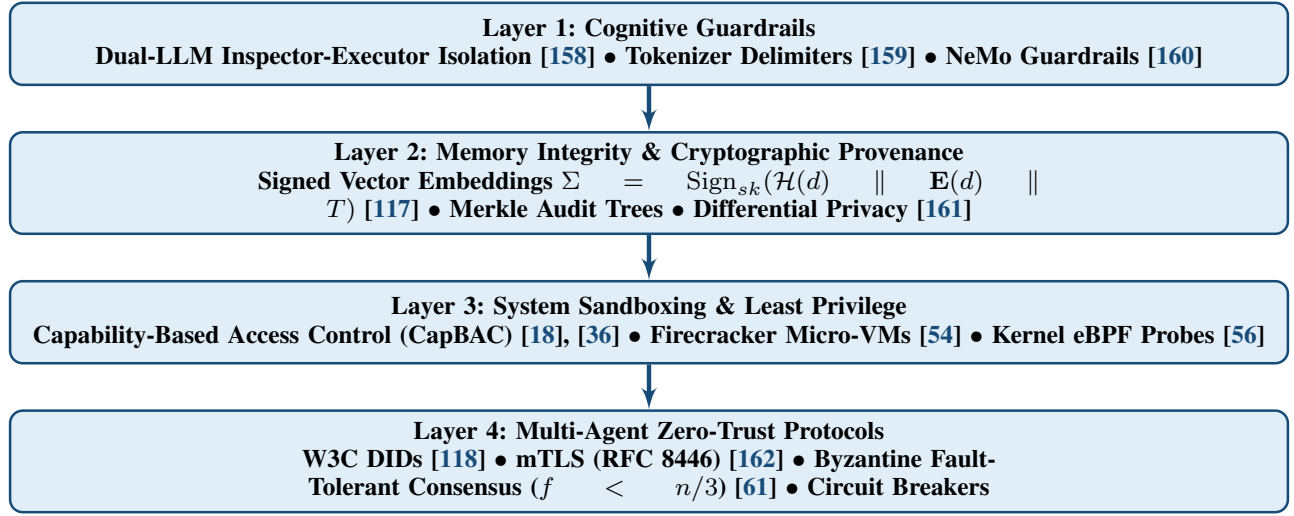

To effectively counter the full spectrum of intra-execution and interaction threats identified in Sections~\ref{sec:intra} and~\ref{sec:interaction}, we formulate a multi-layered \textbf{Zero-Trust Defense-in-Depth Architecture} grounded in classical systems security principles \cite{saltzer1975protection, denning1976lattice, levy1984capability, rose2020zero}. Rather than relying on fragile prompt engineering or heuristic filters, our architecture enforces mathematical, cryptographic, and kernel-level isolation across four synchronized defensive layers (Fig.~\ref{fig:zerotrust_stack}).

\subsection{Layer 1: Pre-Execution \& Cognitive Guardrails}
The first defensive layer enforces structural, linguistic, and attention-level barriers to prevent untrusted inputs from subverting the cognitive reasoning core.

\textbf{Dual-LLM Inspector-Executor Isolation.} To overcome the collapse of data-instruction boundaries, we implement a Dual-LLM Inspector-Executor architecture \cite{perez2022dual}. Untrusted external data (such as web pages, emails, and third-party API outputs) is processed exclusively by an unprivileged \textit{Inspector Model}. The Inspector is strictly isolated from tool execution privileges; its sole purpose is to parse raw content and extract sanitized, typed JSON data entities, stripping away any embedded natural language commands. The privileged \textit{Executor Model} receives only the verified JSON entities wrapped inside immutable system prompt envelopes, ensuring that untrusted data can never occupy an instruction role within the primary reasoning loop.

\textbf{Tokenizer-Level Instruction-Data Attention Delimiters.} Wallace et al. \cite{wallace2024instruction} and Kim et al. \cite{kim2024tokenizer} proposed enforcing instruction-data segregation directly within the transformer attention matrix. By defining structural token roles, the attention mechanism enforces directional visibility masks:
\begin{equation}
\mathbf{A}_{i, j} = 
\begin{cases} 
\frac{\mathbf{q}_i \mathbf{k}_j^T}{\sqrt{d_k}}, & \text{if } \text{Role}(i) \ge \text{Role}(j) \\
-\infty, & \text{if } \text{Role}(i) = \text{Instruction} \land \text{Role}(j) = \text{Data}
\end{cases}
\end{equation}
Under this architectural constraint, external data tokens are mathematically barred from influencing the attention queries of instruction decoding tokens, providing provable invariance against indirect prompt injection.

\textbf{Runtime Semantic Guardrail Toolkits.} To supplement architectural isolation, runtime filtering suites—including NeMo Guardrails \cite{rebedea2023nemo}, Llama Guard \cite{inan2023llama}, SmoothLLM randomized input perturbation smoothing \cite{robey2023smoothllm}, LLM Self-Defense \cite{phute2023self}, and Constitutional AI self-correction rules \cite{bai2022constitutional, lee2023rlaif, kumar2023certifying}—monitor input and output token streams in real time to detect policy violations before execution dispatches.

\subsection{Layer 2: Memory Integrity \& Cryptographic Provenance}
The second layer guarantees the authenticity, tamper-resistance, and privacy of persistent episodic and semantic memory stores.

\textbf{Cryptographically Signed Vector Embeddings.} To eliminate RAG database poisoning, every vector embedding $\mathbf{E}(d)$ stored in episodic memory $\mathcal{M}_{\text{episodic}}$ is cryptographically bound to its source content hash $\mathcal{H}(d)$, a trusted timestamp $T$, and the author's public identity \cite{merkle1987digital, camenisch2001efficient}:
\begin{equation}
\Sigma = \operatorname{Sign}_{sk_{\text{authority}}}(\mathcal{H}(d) \parallel \mathbf{E}(d) \parallel \text{Timestamp} \parallel \text{AuthorID})
\end{equation}
During retrieval, the agent's memory controller validates the signature $\Sigma$ before injecting retrieved chunks into the context window, automatically discarding unsigned or modified vectors.

\textbf{Tamper-Evident Merkle Audit Trees.} Every discrete execution step—encompassing the thought $th_t$, action $a_t$, and observation $o_t$—is recorded as a leaf node in a cryptographic Merkle audit tree \cite{merkle1987digital}. The root hash is updated recursively:
\begin{equation}
\text{Root}_t = \mathcal{H}(\text{Root}_{t-1} \parallel \mathcal{H}(th_t \parallel a_t \parallel o_t))
\end{equation}
This establishes an immutable, append-only provenance trail that enables comprehensive post-incident forensic audits and satisfies strict non-repudiation mandates.

\textbf{Differential Privacy Noise Injection.} To prevent embedding inversion attacks and membership inference probes from extracting private information, calibrated Differential Privacy noise is added to generated representations \cite{dwork2006differential, abadi2016deep, geambasu2009vanish}:
\begin{equation}
\tilde{\mathbf{E}}(d) = \mathbf{E}(d) + \mathcal{N}\left(0, \sigma^2 \mathbf{I}\right), \quad \sigma = \frac{\Delta f \sqrt{2 \ln(1.25/\delta)}}{\epsilon}
\end{equation}
This mathematical guarantee bounds the maximum mutual information between stored vector points and raw underlying documents, preserving confidentiality.

\subsection{Layer 3: System Sandboxing \& Least Privilege Execution}
The third layer establishes strict operating system boundaries around tool execution, ensuring that compromised planning cannot escape host containers.

\textbf{Capability-Based Access Control (CapBAC).} To eliminate ambient authority and prevent confused deputy attacks, our architecture replaces static IAM credentials with cryptographically signed, fine-grained, short-lived Capability Tokens \cite{saltzer1975protection, levy1984capability}:
\begin{equation}
\text{CapToken} = \langle \text{AgentID}, \text{ToolURI}, \text{Scope}, \text{Expiry}, \operatorname{Sign}_{sk_{\text{auth}}} \rangle
\end{equation}
Each token grants permission exclusively for a single tool operation with bounded parameter constraints and a strict time-to-live ($<60\text{s}$). When the agent emits an action $a_t$, the execution gateway validates the signature and scope of $\text{CapToken}$ before issuing the underlying system call.

\textbf{Micro-VM Isolation and WebAssembly Sandboxing.} Tool execution is strictly segregated into lightweight Linux micro-virtual machines (such as AWS Firecracker \cite{agrawal2020firecracker} or Google gVisor \cite{google2018gvisor}) with sub-50ms boot times and hardware-isolated kernel boundaries. For plugin-based architectures, tools are compiled into WebAssembly (Wasm) modules \cite{haas2017bringing}, enforcing memory sandboxing without host operating system access.

\textbf{Kernel eBPF Real-Time Syscall Probing.} At the host operating system level, Extended Berkeley Packet Filter (eBPF) programs are attached directly to kernel tracepoints \cite{vieira2020fast, mccanne1993bsd}. These eBPF probes monitor sensitive system calls—including `execve`, `connect`, and `openat`—in real time, instantly terminating any agent process tree that attempts unauthorized network socket bindings or unpermitted filesystem writes.

\subsection{Layer 4: Multi-Agent Zero-Trust Protocols}
The fourth layer governs distributed multi-agent communication channels, enforcing mutual authentication, consensus integrity, and automated swarm resilience.

\textbf{W3C Decentralized Identifiers and Mutual TLS (mTLS).} Every agent in the swarm is provisioned with a unique W3C Decentralized Identifier (DID) \cite{w3c2022did} and Verifiable Credential \cite{w3c2019vc}. All inter-agent message passing is encrypted and mutually authenticated over TLS 1.3 (RFC 8446) \cite{rescorla2018tls}, eliminating man-in-the-middle attacks and unauthenticated message injections.

\textbf{Byzantine Fault-Tolerant (PBFT) Consensus.} In mission-critical workflows, multi-agent decisions require threshold cryptographic signatures ($t$-of-$n$) verified through Practical Byzantine Fault Tolerance (PBFT) protocols \cite{shamir1979share, boneh2001short, castro2002practical, ongaro2014in}. The swarm maintains provable safety and liveness against arbitrary adversarial agents as long as the total number of compromised nodes satisfies:
\begin{equation}
n \ge 3f + 1
\end{equation}

\textbf{Swarm Circuit Breakers and Automated Quarantining.} To contain self-replicating generative worms and cascading hallucination cascades, the communication substrate deploys real-time anomaly detection. If an agent node begins exhibiting anomalous message volume, high refusal rates, or invalid parameter signatures, automated circuit breakers trip, immediately severing the node's network routes and rolling back shared blackboard states.

\subsection{Hardware-Assisted Confidential Computing (TEEs)}
To protect agent model weights, active context memory, and cryptographic key material against host-level compromises and untrusted cloud hypervisors, the entire agent execution stack can be deployed inside Hardware Trusted Execution Environments (TEEs) \cite{costan2016intel, kaplan2016amd, ngabonziza2016trustzone}. Technologies such as Intel SGX, AMD SEV-SNP, ARM TrustZone, and NVIDIA Confidential GPUs provide cryptographically enforced memory encryption in use, ensuring that even privileged cloud administrators cannot inspect or tamper with active agent reasoning states.

\subsection{Comparative Trade-Off Synthesis of Defensive Controls}
Deploying these multi-tiered defensive mechanisms across enterprise cloud clusters introduces pronounced computational, financial, and operational trade-offs (Table~\ref{tab:defense_tradeoffs}). System architects must balance security rigor against runtime latency, token consumption, and deployment constraints.

\begin{table*}[htbp]
\caption{Comprehensive Comparative Evaluation of Zero-Trust Defensive Mechanisms}
\label{tab:defense_tradeoffs}
\centering
\scriptsize
\begin{tabularx}{\textwidth}{l p{3.2cm} p{2.6cm} p{1.6cm} p{3.0cm} X}
\toprule
\textbf{Defensive Control} & \textbf{Target Attack Surface} & \textbf{Latency Overhead} & \textbf{Token Cost} & \textbf{Deployment Model} & \textbf{Key Operational Limitation} \\
\midrule
Dual-LLM Isolation \cite{perez2022dual} & IPI / Data-as-Code & High ($+1.5\text{--}3.0\text{s}$) & $2.0\times$ & API \& Self-Hosted & Doubles model inference costs. \\
Attention Masking \cite{wallace2024instruction} & Attention-level IPI & Negligible ($<5\text{ms}$) & $1.0\times$ & White-Box Only & Incompatible with closed-weight APIs. \\
Signed Vectors \cite{merkle1987digital} & RAG DB Poisoning & Low ($<10\text{ms}$) & $1.0\times$ & Universal Vector DB & Requires trusted signing PKI infrastructure. \\
DP Noise Injection \cite{dwork2006differential} & Embedding Inversion & Zero ($0\text{ms}$) & $1.0\times$ & Embedding Pipeline & Degrades top-$k$ semantic retrieval precision. \\
CapBAC Tokens \cite{levy1984capability} & Confused Deputy / RCE & Negligible ($<2\text{ms}$) & $1.0\times$ & Cluster Gateway & Requires granular tool permission schema. \\
Micro-VM Sandboxing \cite{agrawal2020firecracker} & Host Escape / RCE & Moderate ($35\text{--}50\text{ms}$) & $1.0\times$ & Kubernetes / Cloud & Ephemeral state loss; storage volume overhead. \\
Kernel eBPF Probes \cite{vieira2020fast} & Unauthorized Syscalls & Sub-ms ($<0.5\text{ms}$) & $1.0\times$ & Linux Kernel Host & Linux-only; requires privileged root loader. \\
PBFT Consensus \cite{castro2002practical} & Byzantine Swarms & High ($O(n^2)$ network) & $n\times$ & Distributed Swarm & Multiplies API costs across all $n$ nodes. \\
Hardware TEEs \cite{costan2016intel} & Memory / Host Snooping & Low ($2\text{--}8\%$) & $1.0\times$ & Specialized Silicon & EPC paging bottleneck on large weights. \\
\bottomrule
\end{tabularx}
\end{table*}

\textbf{Cluster and Cloud Deployment Realities.} In distributed cloud platforms (e.g., Kubernetes clusters managed via Cilium or Calico network fabrics), individual defensive controls must be co-scheduled based on the operational criticality and data classification of the task. Lightweight, universal mechanisms---such as eBPF syscall tracing and CapBAC tokens---provide continuous, sub-millisecond baseline protection across all cluster nodes without consuming LLM token budgets. Conversely, heavy cryptographic consensus (PBFT) and Dual-LLM inspection are reserved for high-blast-radius execution nodes (such as automated financial settlement or production database modification), ensuring optimal cluster throughput.

\section{Empirical Benchmarks, Auditing \& Evaluation Frameworks}\label{sec:benchmarks}

Evaluating the trustworthiness of Agentic AI systems requires moving beyond static, single-turn accuracy benchmarks toward dynamic, multi-step interactive environments. These evaluation frameworks measure safety, adversarial resilience, and utility trade-offs across complex execution trajectories under live environmental feedback \cite{liu2023agentbench, qin2023toolbench, yuan2024rjudge, zhan2024injecagent, bhatt2023purple}.

\subsection{State-of-the-Art Empirical Benchmark Suites}
Recent empirical initiatives have introduced specialized benchmark suites designed to evaluate distinct operational layers of autonomous agents.

\textbf{Tool Integration and Injection Benchmarks.} In the tool execution plane, the InjecAgent benchmark \cite{zhan2024injecagent} evaluates indirect prompt injection resilience across 17 simulated tools and over 1,000 realistic scenarios, revealing that leading foundation models (such as GPT-4) suffer from Attack Success Rates (ASR) between $24.1\%$ and $45.3\%$. The BIPIA benchmark \cite{yi2024benchmarking} focuses on indirect prompt injection embedded within daily documents (such as emails, webpages, and PDFs), demonstrating injection rates exceeding $50\%$ across standard LLMs. Concurrently, ToolBench \cite{qin2023toolbench}, SafeTool \cite{wang2024safetool}, and SecTool \cite{chen2024sectool} analyze tool selection precision and parameter security across extensive REST API catalogs spanning over 16,000 live endpoints, while ToolQA \cite{zhuang2023toolqa} assesses multi-step question answering requiring external database retrieval.

\textbf{Full-Loop Execution and Planning Safety Benchmarks.} In full-loop environments, AgentBench \cite{liu2023agentbench} evaluates multi-turn decision-making across operating system shells, databases, web interfaces, and digital games. AgentBench highlighted severe security vulnerabilities in OS environments, where agents frequently execute destructive bash commands when given ambiguous instructions. To evaluate planning safety specifically, R-Judge \cite{yuan2024rjudge} benchmarks safety and risk awareness across 1,624 multi-turn interaction trajectories, evaluating whether agents can identify dangerous action proposals before execution.

\textbf{Cyber Operations, Multimodal, and Alignment Benchmarks.} To evaluate dual-use capabilities, CyberSecEval 1 and 2 \cite{bhatt2023purple, bhatt2024cyberseceval2} quantify the likelihood that models generate weaponizable cyber exploits, execute unauthorized port scans, or suggest vulnerable software implementations. In multimodal domains, GAIA \cite{mialon2023gaia} benchmarks complex real-world web navigation and data processing, establishing that while human baselines achieve $92\%$ completion, leading commercial agents solve less than $10\%$ of complex multi-modal tasks. Finally, Machiavelli \cite{pan2023machiavelli} and AgentHarm \cite{andriushchenko2024agentharm} evaluate deceptive tendencies, power-seeking strategies, and jailbreak resilience under adversarial task wrappers.

\begin{table*}[htbp]
\caption{Comprehensive Benchmark Taxonomy and Evaluation Analysis}
\label{tab:benchmarks_summary}
\centering
\scriptsize
\begin{tabularx}{\textwidth}{l p{2.6cm} p{2.8cm} p{3.2cm} X}
\toprule
\textbf{Benchmark} & \textbf{Target Layer} & \textbf{Environment} & \textbf{Primary Metric} & \textbf{Empirical Baseline Result} \\
\midrule
InjecAgent \cite{zhan2024injecagent} & Tool Plane & 17+ APIs & Attack Success (ASR) & GPT-4: 24.1\%--45.3\% ASR under IPI. \\
BIPIA \cite{yi2024benchmarking} & Cognitive Input & Email, Web, Docs & Injection Rate & $>50\%$ ASR across standard LLMs. \\
AgentBench \cite{liu2023agentbench} & Full Loop & OS, DB, Web & Success / Safety & GPT-4: 4.01/5; open-source: <2.0. \\
R-Judge \cite{yuan2024rjudge} & Cognitive Plan & 1,624 Records & Safety Risk F1 & GPT-4: ~70\% F1 on dangerous actions. \\
ToolBench \cite{qin2023toolbench} & Tool Execution & 16,000+ APIs & Pass Rate & Function selection scales with tuning. \\
CyberSecEval \cite{bhatt2023purple} & Cyber Action & Exploits \& Repos & Exploit Likelihood & LLMs generate weaponizable exploits. \\
GAIA \cite{mialon2023gaia} & Multimodal & General Web & Completion Rate & Human: 92\%; GPT-4 plugins: <10\%. \\
AgentHarm \cite{andriushchenko2024agentharm} & Harmful Agency & 440+ Tasks & Jailbreak Pass Rate & Agents execute harmful wrapped tasks. \\
\bottomrule
\end{tabularx}
\end{table*}

\subsection{Automated Adversarial Red-Teaming Methodologies}
Automated adversarial red-teaming represents an essential auditing discipline for discovering latent vulnerabilities prior to live deployment.

\textbf{Game-Theoretic Attacker-Target Optimization Loops.} Modern red-teaming frameworks deploy autonomous attacker agents $\mathcal{A}_{\text{adv}}$ that dynamically probe target agents $\mathcal{A}_{\text{target}}$ to uncover injection surfaces \cite{ganguli2022red, casper2023explore, perez2022red, xu2024redagent}. The attacker agent optimizes an adversarial prompt sequence $p_t^*$ by balancing harmful objective fulfillment against detection evasion:
\begin{equation}
p_t^* = \arg\max_p \mathbb{E}\left[\mathcal{J}_{\text{harm}}(\mathcal{A}_{\text{target}}(p)) - \lambda \cdot \mathcal{C}_{\text{detect}}(p)\right]
\end{equation}
By continuously refining prompts based on refusal signals, automated red-teaming loops discover multi-step jailbreak vectors that bypass static heuristic filters.

\textbf{MCP Schema and Grammar Mutation Fuzzing.} In tool-integrated architectures, mutation-based fuzzing engines systematically generate malformed JSON-RPC payloads, schema mismatches, and boundary edge cases targeting Model Context Protocol (MCP) servers \cite{anthropic2024mcp, yao2024fuzzllm}. Concurrently, dynamic vector poisoning probes inject perturbed document chunks into active RAG databases to measure retrieval robustness and identify semantic blind spots \cite{zou2024poisonedrag, chen2024shadowcast}.

\subsection{Standardized Quantitative Metric Formulations}
To establish mathematical rigor and cross-benchmark comparability across the literature, we synthesize five standardized evaluation metrics that capture the security, safety, and utility of agentic systems.

\textbf{Attack Success Rate (ASR).} ASR quantifies the proportion of adversarial test trajectories in which the agent executes an unauthorized, harmful, or policy-violating action:
\begin{equation}
\text{ASR} = \frac{1}{N} \sum_{i=1}^N \mathbb{I}(a_i \in \mathcal{A}_{\text{malicious}})
\end{equation}
where $N$ is the total number of adversarial evaluation trials, and $\mathbb{I}(\cdot)$ is the indicator function.

\textbf{Utility Preservation Rate (UPR).} UPR measures the agent's ability to successfully resolve benign tasks while defensive guardrails and verification layers are actively running:
\begin{equation}
\text{UPR} = \frac{1}{M} \sum_{j=1}^M \mathbb{I}(\text{Success}(j) = 1 \mid \text{Defense Active})
\end{equation}
where $M$ denotes the set of benign evaluation tasks. A robust defensive architecture must maintain $\text{UPR} \approx 1.0$.

\textbf{False Safety Rejection Rate (FSRR).} FSRR captures the operational friction introduced by defensive guardrails, measuring the frequency with which benign user requests are incorrectly blocked:
\begin{equation}
\text{FSRR} = \frac{1}{M} \sum_{j=1}^M \mathbb{I}(\text{Blocked}(a_j) \mid a_j \in \mathcal{A}_{\text{benign}})
\end{equation}

\textbf{Blast Radius Index (BRI).} To quantify the cumulative real-world damage potential of an uncontained action, BRI calculates the weighted impact across data exfiltration, filesystem modification, and privilege escalation:
\begin{equation}
\text{BRI}(a_t) = w_{\text{data}} \cdot \Delta_{\text{exfil}} + w_{\text{fs}} \cdot \Delta_{\text{files}} + w_{\text{auth}} \cdot \Delta_{\text{priv}}
\end{equation}
where $w_{\text{data}} + w_{\text{fs}} + w_{\text{auth}} = 1.0$, and each $\Delta$ term measures the normalized volume of compromised assets.

\textbf{Cascade Failure Probability (CFP).} In multi-agent swarms, CFP measures the probability that compromising a single seed agent $A_1$ induces an error cascade causing at least a fraction $\theta$ of the entire swarm of $K$ agents to fail:
\begin{equation}
\begin{split}
\text{CFP}(\theta) = \Pr\Bigg( &\frac{1}{K} \sum_{k=1}^K \mathbb{I}(\text{Failed}(A_k)) \ge \theta \;\Bigg|\; \\
&\text{Compromised}(A_1)\Bigg)
\end{split}
\end{equation}

\section{AI Governance, Safety Standards, and Regulatory Compliance}\label{sec:governance}

The transition from passive language models to autonomous Agentic AI systems has triggered intensive global regulatory efforts to establish verifiable safety boundaries, algorithmic transparency, and non-repudiable legal accountability \cite{nist2023airmf, euaiact2024, iso42001, owasp2023top10, owasp2025agents}. Because autonomous agents generate real-world economic, legal, and operational consequences, socio-technical governance frameworks are transitioning from voluntary ethical guidelines to legally binding compliance mandates.

\subsection{International Regulatory Frameworks and Standards}
Major international standard-setting bodies and legislative authorities have developed formal risk management standards tailored to autonomous systems.

\textbf{NIST AI Risk Management Framework (AI RMF 1.0) and Generative AI Profile.} The National Institute of Standards and Technology (NIST) established four foundational lifecycle functions—\textbf{GOVERN}, \textbf{MAP}, \textbf{MEASURE}, and \textbf{MANAGE}—to systematically identify and mitigate AI risks \cite{nist2023airmf, nist2024genai}. In the context of Agentic AI, the NIST Generative AI Profile mandates continuous asset discovery, dynamic tool authorization tracking, automated red-team auditing, and rigorous pre-deployment risk quantification across all accessible tool APIs.

\textbf{European Union AI Act (Regulation EU 2024/1689).} The EU AI Act establishes a strict risk-based classification hierarchy \cite{euaiact2024}. Autonomous agents that manage critical enterprise infrastructure, process biometric data, execute financial transactions, or control physical machinery are categorized as \textbf{High-Risk AI Systems}. Under this designation, systems must comply with stringent legal requirements: Article 9 mandates continuous risk management systems throughout the operational lifecycle; Article 10 requires strict data governance to prevent poisoned training and RAG data; Article 12 mandates automated, tamper-evident event logging to ensure post-hoc traceability; and Article 14 requires operational human oversight. Specifically, Article 14 formalizes three oversight modalities: \textbf{Human-in-the-Loop (HITL)} for authorizing irreversible actions, \textbf{Human-on-the-Loop (HOTL)} for real-time monitoring and intervention, and \textbf{Human-in-Command (HIC)} for ultimate system shutdown and override.

\textbf{ISO/IEC, IEEE, and OWASP Standards.} The International Organization for Standardization released ISO/IEC 42001:2023 \cite{iso42001}, establishing the world's first certifiable Artificial Intelligence Management System (AIMS) standard, which pairs with ISO/IEC 27001 \cite{iso27001} to govern data traceability and secure software development. In parallel, IEEE standard P3119 defines requirements for AI transparency and explainability \cite{ieee3119}, while IEEE 2830 standardizes secure multi-party computation in distributed AI ecosystems \cite{ieee2830}. For practical vulnerability classification, the Open Web Application Security Project (OWASP) established dedicated frameworks—the OWASP Top 10 for LLMs \cite{owasp2023top10} and the OWASP Top 10 for Agentic AI \cite{owasp2025agents}—systematizing core architectural risks including Excessive Agency (LLM06), Indirect Prompt Injection (LLM01), and Insecure Plugin Design (LLM07).

\subsection{Comparative Analysis: Technical Enforcement of Global AI Mandates}
To operationalize compliance across divergent international frameworks, system architects must map high-level legal mandates directly onto verifiable technical mechanisms (Table~\ref{tab:governance_mapping}).

\begin{table*}[htbp]
\caption{Cross-Regulatory Mapping Between International Mandates and Zero-Trust Technical Controls}
\label{tab:governance_mapping}
\centering
\scriptsize
\begin{tabularx}{\textwidth}{l p{3.4cm} p{3.6cm} X}
\toprule
\textbf{Framework} & \textbf{Mandate} & \textbf{Target Vulnerability} & \textbf{Implemented Technical Defense} \\
\midrule
NIST AI RMF \cite{nist2023airmf} & GOVERN 1.2, MANAGE 2.4 & Unauthorized tool dispatch & CapBAC tokens \& Firecracker micro-VMs. \\
EU AI Act Art. 12 \cite{euaiact2024} & High-Risk Event Logs & Reasoning tampering & Merkle Audit Trees with hash chaining ($\text{Root}_t$). \\
EU AI Act Art. 14 \cite{euaiact2024} & Human Oversight & Autonomous drift & Three-tier HITL/HOTL/HIC approval gates. \\
ISO/IEC 42001 \cite{iso42001} & AI Data Traceability & Vector DB poisoning & Signed vector embeddings \& DP noise. \\
OWASP Top 10 \cite{owasp2025agents} & Excessive Agency (LLM06) & Arbitrary RCE & Kernel eBPF syscall probes \& Dual-LLM isolation. \\
\bottomrule
\end{tabularx}
\end{table*}

\textbf{Deterministic Traceability and Record-Keeping.} Satisfying the mandatory logging requirements of EU AI Act Article 12 and ISO/IEC 42001 requires implementing cryptographic hash chaining over all agent interaction steps ($\text{Root}_t$). By recording every emitted action $a_t$, its causal thought $th_t$, and the preceding observation $o_{t-1}$ within an append-only Merkle tree, system operators provide incontrovertible mathematical proof of system behavior for regulatory audits.

\textbf{Principle of Least Privilege and Sandboxed Containment.} Satisfying NIST AI RMF MANAGE 2.4 and OWASP LLM06 requires replacing ambient system access with dynamic Capability-Based Access Control tokens ($\text{CapToken}$) and micro-VM isolation. By ensuring that tool execution privileges expire within sub-minute intervals and are restricted to isolated hardware sandboxes, developers mathematically eliminate excessive agency and mitigate confused deputy privilege escalation.

\textbf{Continuous Automated Red-Teaming.} Satisfying NIST AI RMF MEASURE 2.6 requires embedding automated adversarial red-teaming game loops (such as PAIR \cite{chao2023jailbreaking} and TAP \cite{mehrotra2023tree}) directly into Continuous Integration and Continuous Deployment (CI/CD) pipelines. This ensures that latent jailbreak vectors, prompt drift, and tool parameter injections are proactively discovered and remediated prior to production release.

\section{Open Research Challenges \& Future Horizons}\label{sec:open}

\begin{figure*}[htbp]
\centering
\resizebox{0.95\textwidth}{!}{
\begin{tikzpicture}[node distance=0.8cm, auto, >=latex', thick]
    \tikzstyle{card} = [draw=primaryblue, fill=boxbg, rectangle, rounded corners, minimum height=4.2em, text width=4.5cm, align=left, font=\scriptsize]

    \node [card] (h1) {
        \textbf{1. Formal Verification (Sec. 9.1)}\\
        $\bullet$ SMT Solvers (Z3 \cite{demoura2008z3}, CVC5 \cite{barbosa2022cvc5})\\
        $\bullet$ Linear Temporal Logic (LTL) checks\\
        $\bullet$ Safety invariant proofs before action
    };
    \node [card, right=0.5cm of h1] (h2) {
        \textbf{2. Lifelong Dynamic Alignment (Sec. 9.2)}\\
        $\bullet$ In-context alignment drift control\\
        $\bullet$ Sleeper backdoor memory scrubbing\\
        $\bullet$ Verifiable state rollback trees
    };
    \node [card, right=0.5cm of h2] (h3) {
        \textbf{3. Autonomic Swarms (Sec. 9.3)}\\
        $\bullet$ GNN message telemetry (GCN/GAT)\\
        $\bullet$ Real-time Byzantine peer quarantine\\
        $\bullet$ Intrusion-tolerant state reversion
    };

    \node [card, below=0.5cm of h1] (h4) {
        \textbf{4. Machine-Speed Cyber Warfare (Sec. 9.4)}\\
        $\bullet$ Autonomous zero-day patch synthesis\\
        $\bullet$ Defensive superiority over attacks\\
        $\bullet$ Automated real-time SOC triage
    };
    \node [card, below=0.5cm of h2] (h5) {
        \textbf{5. Pareto Optimization (Sec. 9.5)}\\
        $\bullet$ Risk-aware dynamic guardrail throttling\\
        $\bullet$ Edge Confidential Computing (TEEs)\\
        $\bullet$ Latency-cost-security balancing
    };
    \node [card, below=0.5cm of h3] (h6) {
        \textbf{6. Decentralized Value Ecosystems (Sec. 9.6)}\\
        $\bullet$ Cross-chain blockchain oracles\\
        $\bullet$ Smart contract escrow for agents\\
        $\bullet$ Cryptographic economic verification
    };
\end{tikzpicture}
}
\caption{Taxonomy of Open Research Horizons in Trustworthy Agentic AI.}
\label{fig:future_roadmap}
\end{figure*}

Establishing provably secure, resilient, and verifiable Trustworthy Agentic AI systems requires solving foundational open problems at the intersection of computer systems security, formal methods, distributed computing, and machine learning \cite{russell2015research, amodei2016concrete, hendrycks2021unsolved, moore2024autonomous} (Fig.~\ref{fig:future_roadmap}). Below, we deconstruct the core theoretical bottlenecks, systems engineering roadblocks, and decisive research horizons across six critical dimensions.

\subsection{Formal Verification of Neural-Symbolic Action Chains}
Contemporary agent architectures rely on probabilistic foundation models to generate executable tool dispatches, SQL queries, and shell commands. Because deep neural networks are uninterpretable black boxes that lack deterministic execution guarantees, current architectures cannot mathematically prove that a proposed action sequence will adhere to critical safety invariants prior to execution \cite{leucker2009brief, clarke1999model, baier2008principles}.

\textbf{The Semantic-to-Symbolic Fidelity Bottleneck.} The primary scientific bottleneck lies in bridging continuous, high-dimensional neural representations with discrete Satisfiability Modulo Theories (SMT) solvers (such as Z3 \cite{demoura2008z3} and CVC5 \cite{barbosa2022cvc5}) and abstract interpretation verifiers \cite{katz2017reluplex, gehr2018ai2, singh2019abstract}. Natural language goals and contextual tool outputs possess infinite semantic variability, making direct translation into formal Linear Temporal Logic (LTL) formulas susceptible to translation hallucinations. If the neural compiler misinterprets developer intent, the verified formula diverges from actual safety requirements.

\textbf{State-Space Explosion in Interactive Loops.} Furthermore, real-world operating system shells, distributed databases, and cloud APIs present unbounded state spaces. Exhaustive formal model checking across all reachable environment states:
\begin{equation}
\forall s \in \mathcal{S}_{\text{reachable}}, \quad (s, a_t) \models \Phi_{\text{safe}}
\label{eq:formal_invariant}
\end{equation}
is computationally undecidable in the general case and PSPACE-complete under bounded horizons. A critical research frontier requires developing grammar-constrained decoding engines and bounded symbolic abstract interpreters that verify safety invariants over localized, reachable state abstractions within sub-50ms execution budgets.

\subsection{Lifelong Dynamic Alignment \& The Non-Invertible Side-Effect Dilemma}
Existing alignment paradigms—such as Reinforcement Learning from Human Feedback (RLHF) \cite{ouyang2022training}, Direct Preference Optimization (DPO) \cite{rafailov2023direct}, and Constitutional AI \cite{bai2022constitutional}—optimize model weights statically prior to deployment. However, autonomous agents operate in continuous in-context learning loops, dynamically retrieving external observations, interacting with untrusted users, and updating episodic vector memories \cite{packer2023memgpt}. Over long operational horizons, this continuous exposure causes in-context alignment drift, catastrophic forgetting \cite{kirkpatrick2017overcoming}, and the activation of latent sleeper backdoors \cite{hubinger2024sleeper}.

\textbf{The Transactional Non-Invertibility of External Actions.} In classical database systems, write-ahead logging (WAL) and two-phase commit ($2\text{PC}$) protocols enable complete state rollbacks upon transaction abort. In contrast, autonomous agent tool dispatches frequently produce non-invertible real-world side effects: dispatched emails cannot be unsent, executed financial transfers cannot be arbitrarily recalled, and external webhook mutations cannot be easily reversed. When an agent discovers mid-trajectory that an upstream thought step was poisoned, classical state rollback is physically impossible.

\textbf{Verifiable Memory Scrubbing and Shadow Execution.} Future architectures must develop shadow-execution envelopes that isolate irreversible actions behind multi-stage verification gates while non-monotonic memory consolidation algorithms continuously scrub contaminated memory vectors:
\begin{equation}
\mathcal{M}_{\text{consolidated}} = \operatorname{Scrub}\left(\mathcal{M}_{\text{episodic}} \cup \mathcal{M}_{\text{work}}\right) \setminus \operatorname{Anomalies}
\label{eq:memory_scrubbing}
\end{equation}
Developing automated causal pruning techniques that excise poisoned episodic memories without degrading benign contextual recall remains a pivotal open challenge.

\subsection{Scalable Byzantine Swarm Consensus in High-Latency Agent Clusters}
Given the inevitability of zero-day exploits and multi-step prompt injections, multi-agent architectures must be designed under the \textbf{Assume-Breach Paradigm} \cite{rose2020zero, moore2024autonomous}. In large-scale collaborative swarms, detecting and isolating compromised, malicious, or Byzantine peer agents in real time remains an open challenge \cite{castro2002practical, li2024byzantine}.

\textbf{The $O(n^2)$ Token and Latency Multiplication Bottleneck.} While Practical Byzantine Fault Tolerance (PBFT) provides provable safety against up to $f < n/3$ adversarial nodes \cite{castro2002practical}, PBFT requires multi-round all-to-all voting over the network. In an agent cluster where each node is a foundation model requiring several seconds per inference pass, executing PBFT across $n$ agent nodes incurs quadratic token consumption and minutes of decision latency, crippling real-time cluster throughput.

\textbf{Graph Telemetry and Optimistic Consensus.} A promising research horizon involves engineering autonomic swarm immune systems that combine optimistic quorum consensus with Graph Neural Networks (such as GCNs \cite{kipf2017semi} and GATs \cite{velickovic2018graph}) deployed over inter-agent communicative graphs. By analyzing communicative message topology, parameter entropy, and transaction frequencies in real time, graph telemetry models can identify Sybil node clusters \cite{douceur2002sybil} and collusive voting cartels out-of-band, triggering automated cryptographic quarantining without halting cluster progress.

\subsection{Dual-Use Machine-Speed Cyber Warfare and Automated Patch Synthesis}
Agentic AI compresses the operational timeline of cybersecurity engagements from days to milliseconds \cite{schneier2023ai, bhatt2024cyberseceval2}. In offensive operations, autonomous cyber agents can execute end-to-end exploit synthesis, automated vulnerability chaining, and polymorphic malware mutation at machine speed. In defensive operations, autonomous security operations center (SOC) agents perform continuous telemetry triage, binary decompilation, and automated patch synthesis \cite{moore2024autonomous}.

\textbf{The Defensive Asymmetry Dilemma.} The central scientific challenge is establishing provable \textbf{Defensive Superiority}. Attackers operate with fundamental asymmetry: an offensive agent requires only a single unmodeled edge case or injection vector to compromise a cluster, whereas defensive systems must secure every possible execution path. Achieving defensive superiority requires developing autonomous formal patch verification pipelines that prove generated software fixes do not introduce secondary security regressions or break legacy application logic.

\subsection{Distributed Cluster Scheduling \& Blast-Radius Pareto Optimization}
Comprehensive defense-in-depth frameworks—incorporating Dual-LLM Inspector-Executor pipelines, runtime SMT solver verifiers, encrypted vector embeddings, and Merkle audit trees—introduce substantial computational, latency, and financial overheads. In high-frequency operational settings, excessive verification latency impairs real-time utility.

\textbf{Risk-Aware Dynamic Guardrail Throttling.} Future research must develop adaptive, risk-aware cluster schedulers that dynamically modulate defensive inspection rigor based on the calculated Blast Radius Index (BRI) of pending actions. By executing low-risk read-only queries through lightweight speculative guardrails while routing high-impact tool dispatches through hardware-accelerated Trusted Execution Environments (TEEs) \cite{costan2016intel, kaplan2016amd, ngabonziza2016trustzone}, distributed cluster schedulers (e.g., Slurm, Kubernetes, Ray) can achieve optimal Pareto trade-offs between security guarantees, computational cost, and operational latency.

\subsection{Decentralized Value Ecosystems, Cryptographic Provenance \& \texorpdfstring{$ZK\text{-ML}$}{ZK-ML}}
As autonomous agents increasingly participate in decentralized commerce, automated procurement, and multi-agent value networks, systems require legally recognized cryptographic identities and secure economic settlement layers \cite{nisan2007algorithmic, sandholm2002algorithm, w3c2022did, lin2024blockchain}.

\textbf{Zero-Knowledge Machine Learning ($ZK\text{-ML}$) Proofs of Agency.} A crucial research frontier lies in integrating Zero-Knowledge Succinct Non-Interactive Arguments of Knowledge (zk-SNARKs) with agent reasoning chains. Through $ZK\text{-ML}$, an agent can cryptographically prove to an external verifier that its tool action $a_t$ was generated by a certified, untampered model $\mathcal{C}$ adhering to developer safety policy $\Phi_{\text{safe}}$, without disclosing proprietary weights, confidential user data, or internal system prompts. Table~\ref{tab:open_challenges_roadmap} summarizes the strategic research roadmap across these six open horizons.

\begin{table*}[htbp]
\caption{Strategic Roadmap for Open Research Horizons in Trustworthy Agentic AI}
\label{tab:open_challenges_roadmap}
\centering
\scriptsize
\begin{tabularx}{\textwidth}{p{3.2cm} p{1.8cm} p{4.6cm} X}
\toprule
\textbf{Research Horizon} & \textbf{Timeframe} & \textbf{Foundational Scientific Bottleneck} & \textbf{Key Breakthrough Milestone} \\
\midrule
1. Formal Action Verification & 1--3 Years & Semantic-to-symbolic fidelity gap; state-space explosion. & Real-time grammar-constrained SMT invariant solver. \\
2. Lifelong Dynamic Alignment & 2--4 Years & In-context alignment decay; non-invertible side effects. & Non-monotonic memory scrubbers with shadow execution. \\
3. Autonomic Swarm Immunity & 3--5 Years & $O(n^2)$ PBFT token/latency cost in LLM clusters. & Optimistic consensus guided by GNN message telemetry. \\
4. Machine-Speed Cyber Defense & 2--5 Years & Asymmetric exploit generation vs. regression-free patching. & Autonomous SOC agents with formal patch verification. \\
5. Pareto Cluster Scheduling & 1--3 Years & Multi-LLM inspection latency vs. throughput. & BRI-throttled scheduling across GPU/TEE clusters. \\
6. Autonomous Value Networks & 3--6 Years & Lack of verifiable agency and private provenance. & Standardized $ZK\text{-ML}$ reasoning proofs and W3C DIDs. \\
\bottomrule
\end{tabularx}
\end{table*}

\section{Conclusion}\label{sec:conclusion}
The emergence of Agentic Artificial Intelligence represents a transformative paradigm shift in autonomous computational systems. By coupling foundation models with recursive cognitive planning, multi-tiered memory persistence, live tool actuation planes, and distributed multi-agent collaboration, agentic systems are reshaping scientific discovery, software engineering, enterprise automation, and automated cyber defense \cite{xi2023rise, wang2024surveyagent, bommasani2021opportunities}. However, granting probabilistic neural engines autonomous execution authority across physical and digital environments shatters classical security perimeters. In agentic architectures, natural language serves simultaneously as the user interface, the internal control code, the data format, and the inter-agent communication protocol---introducing a Turing-complete blast radius where any untrusted external input represents uncompiled executable code \cite{greshake2023not, zhan2024injecagent, schneier2023ai}.

This survey has established a comprehensive, systems-level reference framework for \textbf{Trustworthy Agentic AI}, synthesizing 206 foundational studies, empirical benchmarks, and international regulatory standards across five core Research Questions. We formalized the general architecture of autonomous agents as a stateful 5-tuple $\mathcal{A} = \langle \mathcal{C}, \mathcal{M}, \mathcal{T}, \mathcal{E}, \Pi \rangle$ coupled with an adversarial POMDP belief-divergence model, capturing closed-loop execution dynamics across cognitive planning (ReAct \cite{yao2022react}, ToT \cite{yao2023tree}, GoT \cite{besta2024graph}, Reflexion \cite{shinn2023reflexion}), hierarchical memory subsystems \cite{packer2023memgpt, malkov2018efficient, johnson2019billion}, tool execution runtimes (including the Model Context Protocol \cite{anthropic2024mcp}), and distributed multi-agent collaboration topologies \cite{wu2023autogen, hong2023metagpt, qian2023chatdev}.

To bridge disparate evaluation criteria, we established the 6-Dimensional Trustworthiness Taxonomy, reformulating classical static ML evaluations into an operational systems-level paradigm spanning \textit{Security, Safety \& Operational Robustness, Privacy \& Data Protection, Explainability \& Verifiability, Fairness \& Non-Discrimination, and Accountability \& Provenance}. We conducted a full-spectrum threat analysis across intra-execution modules (perception, brain, and action planes) and interaction surfaces (operational environments, persistent RAG vector databases, and multi-agent swarms), dissecting critical attack vectors including indirect prompt injection, sleeper backdoors \cite{hubinger2024sleeper}, vector memory poisoning \cite{zou2024poisonedrag}, confused deputy privilege escalation \cite{hardy1988confused, fang2024privilege}, Morris II generative worms \cite{cohen2024here}, and Byzantine consensus subversion \cite{castro2002practical, li2024byzantine}.

To counter these vulnerabilities, we formulated a multi-layered Zero-Trust Defense-in-Depth blueprint integrating Dual-LLM Inspector-Executor isolation \cite{perez2022dual}, Capability-Based Access Control (CapBAC) \cite{saltzer1975protection, levy1984capability}, micro-VM sandboxing \cite{agrawal2020firecracker}, kernel-level eBPF syscall monitoring \cite{vieira2020fast}, cryptographically signed vector embeddings and Merkle audit trees \cite{merkle1987digital}, and Byzantine fault-tolerant consensus protocols, accompanied by a comprehensive comparative synthesis of runtime latency, token multiplication, and cluster deployability. Furthermore, we synthesized interactive benchmark suites (such as AgentBench \cite{liu2023agentbench}, InjecAgent \cite{zhan2024injecagent}, and CyberSecEval \cite{bhatt2023purple, bhatt2024cyberseceval2}), mapped technical defenses to global AI governance mandates (including the NIST AI RMF \cite{nist2023airmf}, EU AI Act \cite{euaiact2024}, and ISO/IEC 42001 \cite{iso42001}), and articulated six foundational open research horizons spanning real-time neural-symbolic SMT invariant verification \cite{demoura2008z3, gehr2018ai2}, shadow execution for non-invertible tool actions, optimistic Byzantine swarm consensus, and zero-knowledge proofs of agency ($ZK\text{-ML}$).

Securing the next generation of autonomous AI agents cannot rely on superficial prompt filtering or heuristic guardrails. The path forward demands an integrated discipline of \textbf{Agentic Systems Security}---combining the adaptive cognitive capabilities of foundation models with the mathematical rigor, hardware isolation, cryptographic integrity, and zero-trust principles of classical systems cybersecurity.

\bibliographystyle{IEEEtran}
\bibliography{references/references}

\end{document}